\pdfoutput=1
\documentclass[11pt]{article}

\usepackage{acl}

\usepackage{times}
\usepackage{latexsym}

\usepackage[T1]{fontenc}
\usepackage[utf8]{inputenc}

\usepackage{microtype}

\usepackage{inconsolata}

\usepackage{graphicx}

\usepackage{hyperref}       % hyperlinks
\usepackage{url}            % simple URL typesetting
\usepackage{booktabs}       % professional-quality tables
\usepackage{amsfonts}       % blackboard math symbols
\usepackage{nicefrac}       % compact symbols for 1/2, etc.
\usepackage{xcolor}         % colors
\usepackage{multirow}
\usepackage{bbding}
\usepackage{arydshln}
\usepackage{amsmath}
\usepackage{csquotes}
\usepackage{colortbl}
\usepackage[most]{tcolorbox}
\usepackage{arydshln}
\usepackage{adjustbox}
\usepackage{enumitem}

\title{On the Structural Memory of LLM Agents}

\author{
    Ruihong Zeng$^{1*}$,
    Jinyuan Fang$^{1*}$,
    Siwei Liu$^{2\dagger}$,
    Zaiqiao Meng$^{1\dagger}$ \\
    $^1$University of Glasgow \ \ $^2$University of Aberdeen \\
    \texttt{\small zengrh3@gmail.com}, \ \ \texttt{\small j.fang.2@research.gla.ac.uk} \\
    \texttt{\small siwei.liu@abdn.ac.uk}, \ \ \texttt{\small zaiqiao.meng@glasgow.ac.uk}
 }

\begin{document}
\maketitle

\begin{abstract}

\def\thefootnote{*}\footnotetext{Equal contribution.}
\def\thefootnote{$\dagger$}\footnotetext{Corresponding author.}
\def\thefootnote{\arabic{footnote}}

Memory plays a pivotal role in enabling large language model~(LLM)-based agents to engage in complex and long-term interactions, such as question answering (QA) and dialogue systems. While various memory modules have been proposed for these tasks, the impact of different memory structures across tasks remains insufficiently explored. This paper investigates how memory structures and memory retrieval methods affect the performance of LLM-based agents. Specifically, we evaluate four types of memory structures, including chunks, knowledge triples, atomic facts, and summaries, along with mixed memory that combines these components. In addition, we evaluate three widely used memory retrieval methods: single-step retrieval, reranking, and iterative retrieval.  Extensive experiments conducted across four tasks and six datasets yield the following key insights:  (1) Different memory structures offer distinct advantages, enabling them to be tailored to specific tasks;  (2) Mixed memory structures demonstrate remarkable resilience in noisy environments; (3) Iterative retrieval consistently outperforms other methods across various scenarios. Our investigation aims to inspire further research into the design of memory systems for LLM-based agents.~\footnote{All code and datasets are publicly available at:
% \url{https://anonymous.4open.science/r/StructuralMemory-23D1}
\url{https://github.com/zengrh3/StructuralMemory}
}

\end{abstract}

\section{Introduction}
% LLM ==> LLM-based Agents ==> Memory is important ==> Memory components, including memory structures and memory retrieval ==> Cannot identify the impact of memory structures ==> We solve this.  
Large Language Models (LLMs)~\cite{minaee2024large} have attracted widespread attention in natural language tasks due to their remarkable capability.
Recent advancements have significantly accelerated the development of LLM-based agents, with research primarily focusing on profile~\cite{park2023generative, hongmetagpt}, planning~\cite{qian2024chatdev, qiao2024autoact}, action~\cite{qin2023toolllm, wang2024rcagent}, self-evolving~\cite{zhang2024agent} and memory~\cite{packer2023memgpt, lee2024human}.
% Due to their strong performance, LLM-based agents have been widely adopted in various multi-hop question answering~(QA) tasks~\cite{li2024graphreader, lee2024human}.
These innovations have unlocked a wide range of applications across diverse applications~\cite{li2023metaagents, wang2024incharacter, chen2024persona}.
% These advancements have significantly propelled the development of LLM-based agents and has opened up a broad range of applications across various fields, including role-playing~\cite{wang2024incharacter, chen2024persona}, open-world gaming~\cite{wang2023voyager, wangjarvis}, social simulations~\cite{li2023metaagents, park2023generative}, code generation~\cite{ishibashi2024self, bouzenia2024repairagent}, personal assistants~\cite{hongmetagpt, qian2024chatdev, wu2024autogen}, robotics~\cite{sun2024prompt, graule2024gg, zhang2024towards}, and medicine healthcare~\cite{li2024agent, yang2024zhongjing}.

% Just as the human brain relies on memory systems to harness prior experiences for strategy formulation and decision making~\cite{simon1971human, anderson2013architecture, xi2023rise}, agents necessitate specific memory mechanisms to revisit and antecedent strategies to tackle complex reasoning tasks~(e.g., multi-hop QA).
A fundamental element that underpins the effectiveness of LLM-based agents is the memory module.
In cognitive science~\cite{simon1971human, anderson2013architecture}, memory is the cornerstone of human cognition, enabling the storage, retrieval, and drawing from past experiences for strategic thinking and decision-making. 
Similarly, the memory module is vital for LLM-based agents by facilitating the retention and organization of past interactions, supporting complex reasoning capabilities, e.g., multi-hop question answering~(QA)~\cite{li2024graphreader, lee2024human}, and ensuring consistency and continuity in user interactions~\cite{nuxoll2007extending}. 
% Without a robust memory mechanism, these agents struggle to maintain conversational continuity and effectively tackle complex, multi-step reasoning in question answering tasks.
% A fundamental element that underpins the effectiveness of LLM-based agents is the memory module.
% Similar to the human brain, which relies on memory systems to draw from past experiences for strategic thinking and decision-making~\cite{simon1971human, anderson2013architecture, xi2023rise}, the memory module enables agents to retain and organize prior interactions.
% This capability is essential for complex reasoning tasks, such as multi-hop QA, where agents must integrate multiple pieces of information to generate contextually relevant responses.

\begin{figure}
    \centering
    \includegraphics[width=0.99\columnwidth]{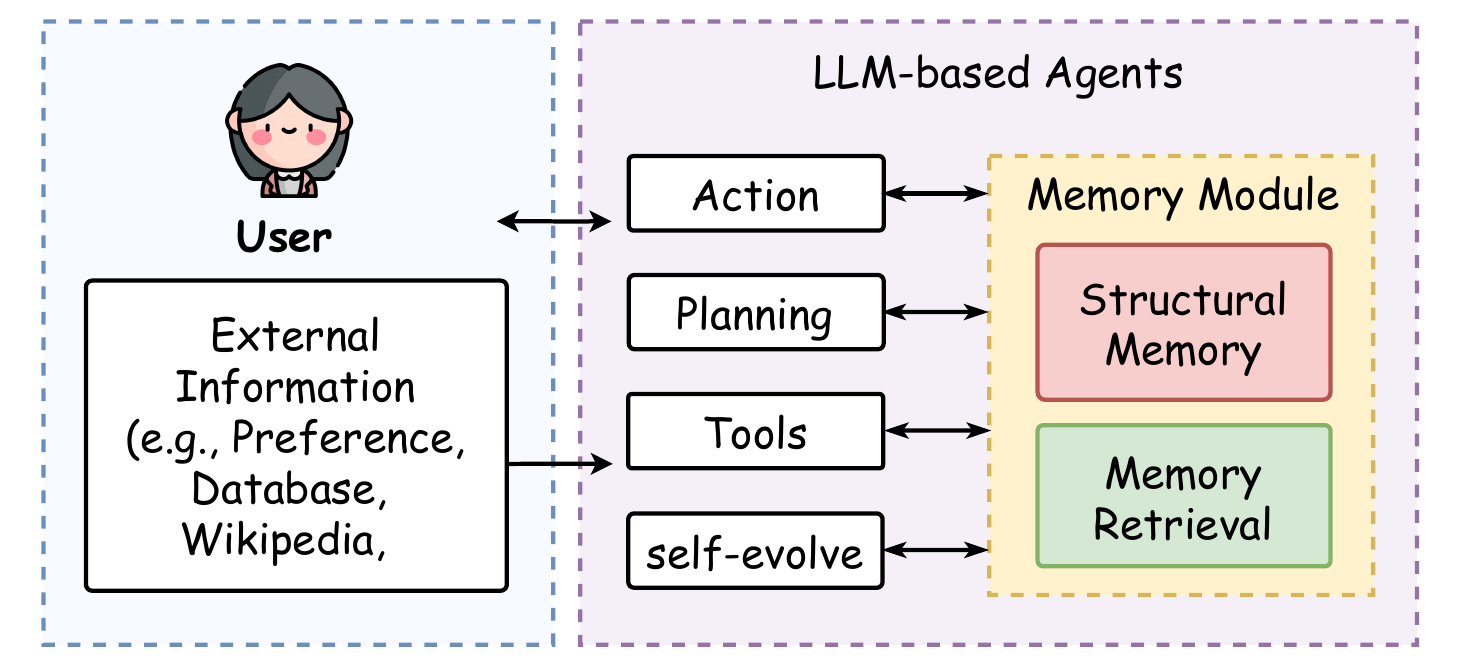}
    \caption{The framework of LLM-based agents, where we focus on the study of memory modules, including memory structures and retrieval methods.}
    \label{fig:trigger}
    \vspace{-1.5em}
\end{figure}

Developing an effective memory module in LLM-based agents typically involves two critical components: structural memory generation and memory retrieval methods~\cite{wang2024survey, zhang2024survey}.
Among the various memory structures used by agents, chunks~\cite{hu2024hiagent}, knowledge triples~\cite{anokhin2024arigraph}, atomic facts~\cite{li2024graphreader}, and summaries~\cite{lee2024human} are the most prevalent.
For instance, HiAgent~\citep{hu2024hiagent} utilizes sub-goals as memory chunks to manage the working memory of LLM-based agents,  ensuring task continuity and coherence,
while Arigraph~\cite{anokhin2024arigraph} adopts knowledge triples, which combine both semantic and episodic memories to store factual and detailed information, making it suitable for complex reasoning tasks.
Meanwhile, ReadAgent~\cite{li2024graphreader} compresses memory episodes into gits memory with summaries manner, organizing them within a structured memory directory.
Upon reviewing the aforementioned memory structures, an important but under-explored question arises: 
\textit{Which memory structures are best suited for specific tasks, and how do their distinct characteristics impact the performance of LLM-based agents?}
This question mirrors how humans organize memory into distinct forms, such as episodic memory for recalling events and semantic memory for understanding relationships~\cite{simon1971human, anderson2013architecture}.
Each form serves a unique purpose, enabling humans to tackle a variety of challenges with flexibility and precision.
Moreover, humans rely on effective retrieval processes to access relevant memories, ensuring the accurate recall of past experiences for problem-solving.
This highlights the need to jointly explore memory structures and retrieval methods to enhance the reasoning capabilities and overall effectiveness of LLM-based agents.
% This mechanism highlights the potential benefits of designing memory structures that are optimally aligned with the demands of specific tasks, thereby enhancing the reasoning capabilities and overall performance of LLM-based agents.

%For instance, given the tasks of multi-hop question answering and dialogue understanding, how do the same memory structures affect performance differently? 
%This raises another critical issue: is it more effective to rely on a one-size-fits-all memory structure, or should multiple structures be tailored and combinedUpon reviewing the aforementioned memory structures, an important but under-explored question arises: How do different memory structures and retrieval methods affect the performance of LLM-based agents across different tasks? 
%These questions stem from the complexity of human memory~\cite{simon1971human, anderson2013architecture}, which integrates both granular details and broader abstractions, enabling flexible recall and reasoning across a wide range of contexts. This complexity highlights the potential benefits of organizing memory in a way that maximizes agent performance, selecting the most appropriate structure to suit the demands of each specific task.

To bridge this gap, we systematically explore the impact of various memory structures and retrieval methods in LLM-based agents.
Specifically, we evaluate existing four types of memory structures: \textit{chunks}~\cite{hu2024hiagent}, \textit{knowledge triples}~\cite{anokhin2024arigraph}, \textit{atomic facts}~\cite{li2024graphreader}, and \textit{summaries}~\cite{li2024graphreader}. 
% Chunks, inspired by episodic memory, store continuous and information-rich sequences to maintain coherence across multiple interactions.
% Knowledge triples, derived from semantic memory, capture relational facts that facilitate logical reasoning. 
% Atomic facts, akin to procedural memory, preserve fine-grained details crucial for precise decision-making.
% Summaries, on the other hand, provide abstract representations by condensing complex information into concise forms, thus supporting efficient reasoning.
% Building on these, we explore \textit{mixed} memory structures that integrate these representations to leverage their complementary strengths for improved performance.
Building on these, we explore the potential of \textit{mixed} memory structures, which combine multiple types of memories to examine whether their complementary characteristics can enhance performance.
Additionally, we assess the robustness of these memory structures to noise, as understanding their reliability under such conditions is essential for ensuring effectiveness across diverse tasks.
Furthermore, we investigate three memory retrieval methods, including \textit{single-step retrieval}~\cite{packer2023memgpt}, \textit{reranking}~\cite{gao2023precise}, and \textit{iterative retrieval}~\cite{li2024corpuslm}, to uncover how different combinations of retrieval methods and memory structures influence overall performance.
% We conduct experiments on four QA tasks: multi-hop QA, single-hop QA, dialogue understanding, and reading comprehension. 
% Our findings indicate that:
% (1) Mixed memories consistently deliver balanced and competitive performance. While chunks and summaries excel in tasks involving extensive and lengthy contexts, i.e., reading comprehension and dialogue understanding, knowledge triples and atomic facts are particularly effective in capturing relational and concise information, making them highly suitable for both multi-hop and single-hop QA; 
% (2) Mixed memory demonstrates remarkable resilience to noise.
% (3) Iterative retrieval stands out as the most effective memory retrieval mechanism across most tasks, such as multi-hop QA, dialogue understanding and reading comprehension.
% , each offering distinct advantages in prioritizing and refining retrieved information, directly influencing the agent's ability to tackle complex tasks.
% These mechanisms differ in their approach to prioritizing and refining the information retrieved, directly influencing the agent's ability to tackle complex question answering tasks.

The main contributions of this work can be summarized as follows: 
% (1) We present the first comprehensive study on the impact of memory structures in LLM-based agents, evaluating four memory structures, e.g., chunks, knowledge triples, atomic facts, and summaries, introducing a mixed memory structure to leverage their strengths, and analyzing their robustness to noise.
% (2) We investigate three retrieval mechanisms: single-step retrieval, reranking, and iterative retrieval, and reveal how these mechanisms interact with memory structures to influence task performance.
% (3) We conduct extensive experiments across four tasks, including multi-hop QA, single-hop QA, dialogue understanding, and reading comprehension.
% All codes and datasets are made publicly available to facilitate reproducibility and further research.
% Notably, mixed memory structures strike an effective balance between comprehensive understanding and precision, even in noisy environments. 
% Moreover, iterative retrieval emerges as the most effective mechanism, consistently delivering superior results across tasks.
(1) We present the first comprehensive study on the impact of memory structures and memory retrieval methods in LLM-based agents on six datasets across four tasks: multi-hop QA, single-hop QA, dialogue understanding, and reading comprehension. 
(2) Our findings reveal that mixed memory consistently achieves balanced and competitive performance across diverse tasks. 
Chunks and summaries excel in tasks involving extensive and lengthy context (e.g., reading comprehension and dialogue understanding), while knowledge triples and atomic facts are particularly effective for relational reasoning and precision in multi-hop and single-hop QA.
Additionally, mixed memory demonstrates remarkable resilience to noise.
(3) Iterative retrieval stands out as the most effective memory retrieval method across most tasks, such as multi-hop QA, dialogue understanding and reading comprehension.

% (1) We comprehensively examine four types of memory structures: chunks, knowledge triples, atomic facts, and summaries, and introduce a mixed memory structure that integrates these types to combine their strengths. 
% Our findings show that tailoring memory structures to task requirements significantly improves performance, with mixed memory excelling in balancing context comprehension and precision, even in noisy environments.
% (2) We analyze three retrieval strategies: single-step retrieval, reranking, and iterative retrieval, demonstrating that iterative retrieval consistently delivers superior results across tasks.
% (3) We conduct the first comprehensive study to investigate the xx of memory structures and memory retrieval mechanisms and identify the main contributing factors.
% (3) We make all associated codes and datasets publicly available to encourage reproducibility and support further research.

\begin{figure*}[!t]
    \centering
    \includegraphics[width=0.99\textwidth]{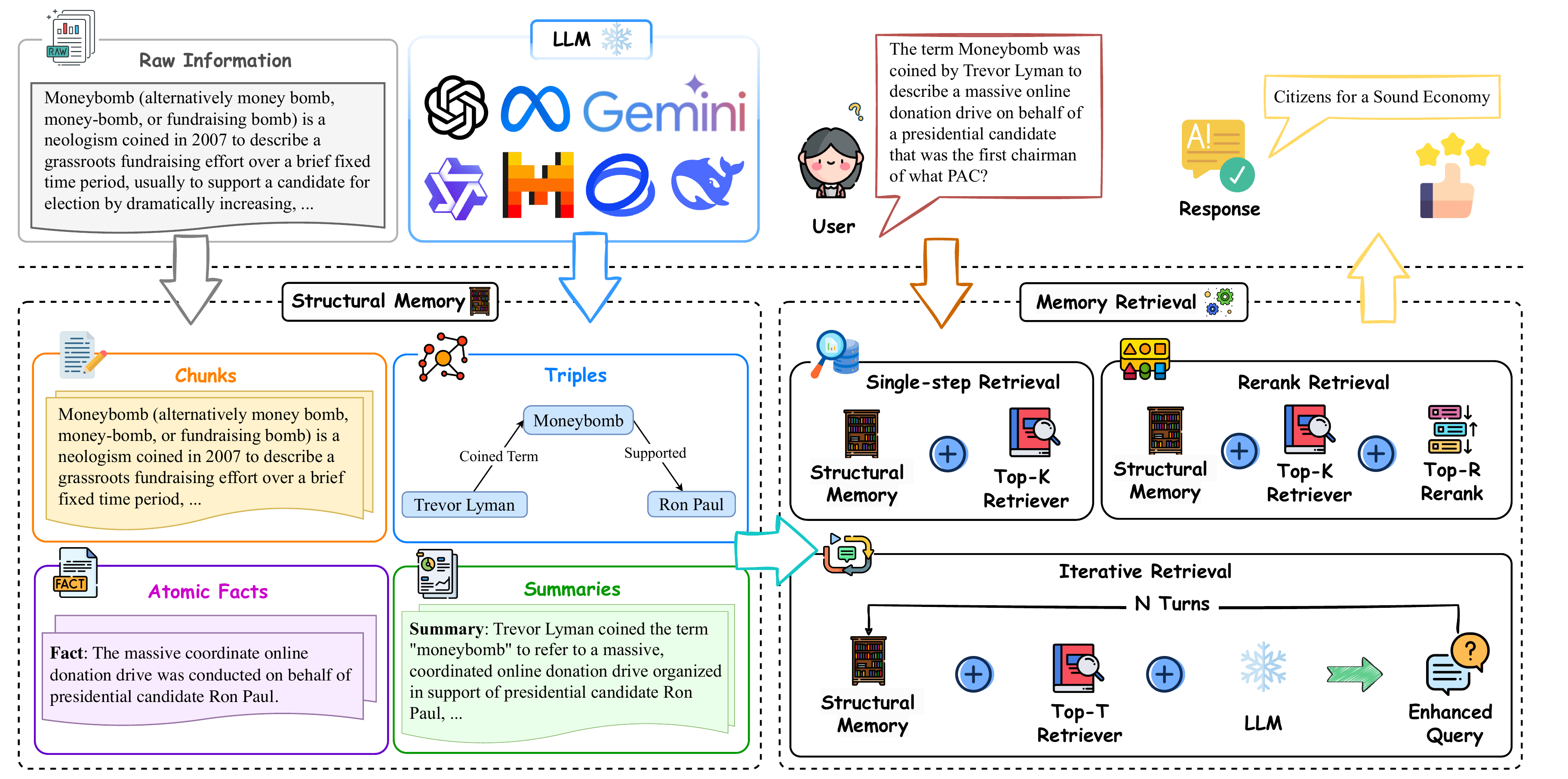}
    \caption{Overview of the memory module workflow in LLM-based agents. Raw information is organized into structural memories, which are processed through retrieval methods to identify the most relevant memories for the query, enabling the generation of precise and contextually enriched responses.}
    \label{fig:overall-process}
    \vspace{-0.5em}
\end{figure*}

\section{Related Works}
\label{sec:related_work}

\subsection{LLM-based Agents}
% So in this section, what we want is to illustrate the basic components of llm-based agents, like memory, action, planning, self-evolving and something like that. 
% Then we would like emphasize the importance of memory module, which we aim to introduce in the next section.
The advent of Large Language Model~(LLM) has positioned them as a transformative step towards achieving Artificial General Intelligence~(AGI)~\cite{wang2024survey}, offering robust capabilities for the development of LLM-based agents~\cite{xi2023rise, xu2024generate}.
% These agents are designed to interact with the external world through action-observation,
Current research in this field primarily focuses on agent planning~\cite{wang2023voyager, yao2024tree, qian2024chatdev, qiao2024autoact}, reflection mechanisms~\cite{shinn2024reflexion, zhang2024agent}, external tools utilization~\cite{qin2023toolllm, wang2024rcagent}, self-evolving capabilities~\cite{zhang2024agent} and memory modules~\cite{hu2024hiagent, lee2024human}.
% These advancements enable agents to exhibit sophisticated reasoning abilities across diverse and challenging scenarios~\cite{wang2024incharacter, hongmetagpt, sun2024prompt, li2024agent}.
% , including role-playing~\cite{wang2024incharacter, chen2024persona}, open-world gaming~\cite{wang2023voyager, wangjarvis}, social simulations~\cite{li2023metaagents, park2023generative}, code generation~\cite{ishibashi2024self, bouzenia2024repairagent}, personal assistants~\cite{hongmetagpt, qian2024chatdev, wu2024autogen}, robotics~\cite{sun2024prompt, graule2024gg, zhang2024towards}, and medicine healthcare~\cite{li2024agent, yang2024zhongjing}.

\subsection{Memory Structures}
Memory module serves as the foundation of LLM-based agents, enabling them to structure knowledge, retrieve relevant information, and leverage prior experiences for reasoning tasks~\cite{zhang2024survey}.
% An effective memory module relies on two key components: the memory structure and the memory retrieval method.
Among the widely adopted memory structures of memory module are chunks~\cite{packer2023memgpt, liu2023think, hu2024hiagent}, knowledge triples~\cite{anokhin2024arigraph}, atomic facts~\cite{li2024graphreader}, and summaries~\cite{lee2024human}.
For instance, HiAgent~\citep{hu2024hiagent} incorporates sub-goals as memory chunks to maintain task continuity and coherence across interactions.
% Similarly, Arigraph~\cite{anokhin2024arigraph} utilizes knowledge triples that integrate semantic and episodic memories, making it well-suited for tasks requiring complex reasoning and logical connections.
On the other hand, GraphReader~\cite{li2024graphreader} employs atomic facts to compress chunks into finer details, providing agents with highly granular information that improves precision in multi-hop question answering tasks.
% Furthermore, ReadAgent~\cite{li2024graphreader} consolidates memory episodes into summaries, organizing them within a structured memory directory.
In this paper, we investigate how various memory structures impact the performance of LLM-based agents. 

\subsection{Memory Retrieval}
The memory retrieval method is another critical component of the memory module, enabling LLM-based agents to retrieve relevant memories to advanced reasoning.
To facilitate this, LLM-based agents often employ retrieval-augmented generation~(RAG)~\cite{lewis2020retrieval, fang-etal-2024-trace}, where relevant memories are first retrieved and then used to generate answers with LLMs. 
In this setting, the retrieved memories are prepended to the queries and serve as input to the LLM to generate response~\cite{ram2023context}.
The most straightforward retrieval method is the single-step retrieval~\cite{packer2023memgpt, zhong2024memorybank}, which aims to identify the Top-$K$ most relevant memories for the query.
% To better align memory retrieval with the specific demands of different tasks, several advancements to the RAG framework have been proposed.
Additionally, reranking~\cite{gao2023precise, ji2024dynamic} leverages the language understanding capabilities of LLMs to prioritize retrieved memories, while iterative retrieval~\cite{li2024corpuslm, shi2024generate} focuses on reformulating queries to improve retrieval accuracy. 
These innovations make memory retrieval more adaptive and consistent with the query, maintaining effective performance across diverse and complex tasks.
In this paper, we explore how different combinations of retrieval methods and memory structures influence overall performance.

\section{Methodology}
% In this section, we begin by introducing the task we aim to explore in Section~\ref{subsec:question_answering}. Next, we present the design of structural memory in Section~\ref{subsec:memory_structure}, followed by a detailed discussion of various memory retrieval mechanisms in Section~\ref{subsec:memory_retrieval_mechanism}.
Figure~\ref{fig:overall-process} illustrates the overview of the memory module within LLM-based agents, highlighting three key components: \textbf{Structural Memory Generation}, \textbf{Memory Retrieval Methods} and \textbf{Answer Generation}.
This section begins with an introduction to structural memory generation in $\S$~\ref{subsec:memory_structure}.
Next, we introduce memory retrieval methods in $\S$~\ref{subsec:memory_retrieval_methods}.
Finally, $\S$~\ref{subsec:answer_generations} discusses answer generation methods.

\subsection{Structural Memory Generation}
\label{subsec:memory_structure}
Structural memory generation enables agents to organize raw documents into structured representations.
By transforming unstructured documents $\mathcal{D}_q$ into structural memory $\mathcal{M}_q$, the agent gains the ability to store, retrieve, and reason over information more effectively.
In this work, we explore four distinct forms of structural memory: chunks $\mathcal{C}_q$, knowledge triples $\mathcal{T}_q$, atomic facts $\mathcal{A}_q$, or summaries $\mathcal{S}_q$. 
% Each form provides a distinct level of granularity, allowing the agent to select the most suitable representation for specific tasks.
% Each form captures information at a unique level of granularity, allowing the selection of the most suitable structure based on the demands of the task.
The generation process for each structural memory is detailed as follows:

\noindent \textbf{Chunks} ($\mathcal{C}_q$).
Chunks~\cite{gao2023retrieval} are a widely used form of structural memory in LLM-based agents.
Each chunk represents a continuous segment of text from a document, typically constrained to a fixed number of tokens $L$.
% , designed to store and access information with continuity and completeness.
% Each chunk corresponds to a fixed-length portion of a document, preserving the natural flow and context of the text. 
% Chunks represent a basic form of structural memory within LLM-based agents and ensure that information is stored and accessed in a manner that remains both continuous and comprehensive.
% This approach ensures that information is stored and accessed in a manner that remains both continuous and comprehensive.
Formally, raw documents $\mathcal{D}_q$ can be divided into a series of chunks, as defined: $\mathcal{C}_q(\mathcal{D}_q) = \{c_{1}, c_{2}, \dots, c_{j}\}$, where each chunk $c_j$ contains at most $L$ tokens.
% \begin{align}
%     p(\mathcal{M}_q | \mathcal{D}_q) &= \mathcal{C}(\mathcal{D}_q) \, , \\
%     \mathrm{where} \quad \mathcal{C}(\mathcal{D}_q) &= \{c_{1}, c_{2}, \dots, c_{j}\} \, .
% \end{align}
% \begin{align}
%     % p(\mathcal{M}_q | \mathcal{D}_q) &= \mathcal{C}(\mathcal{D}_q) \, , \\
%     % \mathrm{where} \quad \mathcal{C}(\mathcal{D}_q) &= \{c_{1}, c_{2}, \dots, c_{j}\} \, .
%     \mathcal{C}_q(\mathcal{D}_q) = \{c_{1}, c_{2}, \dots, c_{j}\} \, ,
% \end{align}
% where each chunk $c_j$ contains up to $L$ tokens.
% This representation of structural memory not only supports the seamless flow of information but also enhances coherence by dividing raw documents into smaller and more manageable segments
% This representation ensures that information is stored and accessed in a manner that remains both continuous and comprehensive.
% ~\cite{packer2023memgpt, liu2023think, hu2024hiagent}.

\begin{tcolorbox}[colback=gray!5!white,colframe=gray!60!black,title=Chunks]
{
\textbf{Definition:} Chunks are continuous, fixed-length segments of text from the document. \\
\textbf{Example:} Generated chunks $\mathcal{C}_q$:
% \textbf{Example:} Assume the document $\mathcal{D}_q$ is: 
% \textit{Moneybomb (alternatively money bomb, money-bomb, or fundraising bomb) is a neologism coined in 2007 to describe a grassroots fundraising effort over a brief fixed time period.}

(1)~\textit{Moneybomb (alternatively money bomb, money-bomb, or fundraising bomb) is a neologism coined in 2007};

(2)~\textit{to describe a grassroots fundraising effort over a brief fixed time period.}
}
\end{tcolorbox}

\noindent \textbf{Knowledge Triples} ($\mathcal{T}_q$).
% Knowledge triples serve as relational structural memories, capturing semantic relationships within documents.
Knowledge triples represent a structured form of memory that captures semantic relationships between entities. 
Each triple is composed of three components: a \textit{head} entity, a \textit{relation}, and a \textit{tail} entity, represented in the format $\langle \text{\textit{head}}; \text{\textit{relation}}; \text{\textit{tail entity}} \rangle$.
Following previous works~\cite{anokhin2024arigraph, fang-etal-2024-trace}, raw documents $\mathcal{D}_q$ are processed by an LLM guided by a tailored prompt $\mathcal{P}_{\mathcal{T}}$ to generate a set of semantic triples $\mathcal{T}_q$.
The generation process can be formally defined as: $\mathcal{T}_q = \texttt{LLM}(\mathcal{D}_q, \mathcal{P}_{\mathcal{T}})$.
% \begin{align}
%     p(\mathcal{M}_q | \mathcal{D}_q) &= p_{\text{LLM}}(\mathcal{T}_q | \mathcal{D}_q, \mathcal{P}_{\mathcal{T}}) \, .
% \end{align}
% \begin{align}
%     \mathcal{T}_q &= \texttt{LLM}(\mathcal{D}_q, \mathcal{P}_{\mathcal{T}}) \, .
% \end{align}
% By structuring knowledge in this way, agents are enabled to perform explicit relational reasoning between entities, which greatly enhances task performance.
% ~\cite{anokhin2024arigraph}. 
% Compared with raw documents that often contain multiple pieces of information, triples offer a finer-grained and more concise representation of memory.
% Each triple encapsulates a single piece of factual information, providing clarity and precision that is particularly beneficial when identifying supporting memories~\cite{anokhin2024arigraph}.

% \begin{tcolorbox}[colback=gray!5!white,colframe=gray!60!black,title=Knowledge Triples]
% \textbf{Definition:} Knowledge triples capture relationships between entities in the form of $\langle$head, relation, tail$\rangle$. \\
% \textbf{Example:} Extracted triples from the text:
% \begin{itemize}
%     \item $\langle$Moneybomb, alternative names, money bomb$\rangle$
%     \item $\langle$Moneybomb, coined in, 2007$\rangle$
%     \item $\langle$Moneybomb, usual purpose, support a candidate for election$\rangle$
% \end{itemize}
% \end{tcolorbox}
\begin{tcolorbox}[
    colback=gray!5!white,
    colframe=gray!60!black,
    title=Knowledge Triples,
]
\textbf{Definition:} Knowledge triples capture relationships between entities. \\
\textbf{Example:} Generated triples $\mathcal{T}_q$:

(1)~$\langle$\textit{Moneybomb; type; neologism}$\rangle$;

(2)~$\langle$\textit{Moneybomb; coined in; 2007}$\rangle$.

% (3)~$\langle$\textit{Moneybomb, usual purpose, support a candidate for election}$\rangle$.
\end{tcolorbox}

\noindent \textbf{Atomic Facts} ($\mathcal{A}_q$).
Atomic facts are the smallest, indivisible units of information, presented as concise sentences that capture essential details. They represent a granular form of structural memory, simplifying raw documents by preserving critical entities, actions, and attributes.
% These facts focus on pivotal nouns, verbs critical to the narrative, and the smallest indivisible units of information, often presented as concise sentences. 
Following~\citet{li2024graphreader}, atomic facts are generated from raw documents $\mathcal{D}_q$ using an LLM guided by a tailored prompt $\mathcal{P}_{\mathcal{A}}$, formally denoted as: $\mathcal{A}_q = \texttt{LLM}(\mathcal{D}_q, \mathcal{P}_{\mathcal{A}})$.
% \begin{align}
%    p(\mathcal{M}_q | \mathcal{D}_q) &= p_{\text{LLM}}(\mathcal{A}_q | \mathcal{D}_q, \mathcal{P}_{\mathcal{A}}) \, .
% \end{align}
% \begin{align}
%    \mathcal{A}_q = \texttt{LLM}(\mathcal{D}_q, \mathcal{P}_{\mathcal{A}}) \, .
% \end{align}
% By structuring memory in this way, atomic facts condense raw documents into a core and compact format, allowing agents to retain only the most essential information while minimizing irrelevance.
% ~\cite{li2024graphreader}.
\begin{tcolorbox}[colback=gray!5!white,colframe=gray!60!black,title=Atomic Facts]
\textbf{Definition:} Atomic facts are the smallest units of indivisible information. \\
\textbf{Example:} Generated atomic facts $\mathcal{A}_q$:

(1)~\textit{Moneybomb is also known as money bomb, money-bomb, or fundraising bomb};

(2)~\textit{Moneybomb is a neologism}.

% (3)~\textit{Moneybomb refers to a grassroots fundraising effort over a brief fixed time period}.

\end{tcolorbox}

\noindent \textbf{Summaries} ($\mathcal{S}_q$).
Summaries provide a condensed and comprehensive description of documents, capturing both global content and key details.
% , summaries enable agents to reason effectively, particularly in tasks involving extensive or complex contexts.
% By compressing the raw document, summaries provide agents with high-level overviews, allowing for effective comprehension and understanding without the need to analyze the entire content.
% Following~\cite{lee2024human}, For raw document $\mathcal{D}_q$, summaries are generated using an LLM guided by prompt $\mathcal{P}_{\mathcal{S}}$, as formally defined below:
Following~\citet{lee2024human}, summaries are generated from raw documents $\mathcal{D}_q$ using an LLM guided by a tailored prompt $\mathcal{P}_{\mathcal{S}}$, defined as: $\mathcal{S}_q = \texttt{LLM}(\mathcal{D}_q, \mathcal{P}_{\mathcal{S}})$.
% \begin{align}
%   p(\mathcal{M}_q | \mathcal{D}_q) &= p_{\text{LLM}}(\mathcal{S}_q | \mathcal{D}_q, \mathcal{P}_{\mathcal{S}}) \, .
% \end{align}
% \begin{align}
%   \mathcal{S}_q = \texttt{LLM}(\mathcal{D}_q, \mathcal{P}_{\mathcal{S}}) \, .
% \end{align}
% By structuring raw documents into summaries-based memories, agents can efficiently capture both global content and granular details, ensuring effective reasoning even in scenarios involving lengthy and complex documents.
% ~\cite{lee2024human}. 
\begin{tcolorbox}[colback=gray!5!white,colframe=gray!60!black,title=Summaries]
\textbf{Definition:} Summaries compress the document into a comprehensive description. \\
\textbf{Example:} Generated summaries $\mathcal{S}_q$:

\textit{Moneybomb, alternatively referred to as money bomb, money-bomb, or fundraising bomb, is a neologism coined in 2007. It describes a grassroots fundraising effort that occurs over a brief fixed time period. 
% This method is typically employed to support a candidate for election by dramatically increasing, concentrating, and publicizing fundraising activities during a specific timeframe.
}
\end{tcolorbox}

\noindent \textbf{Mixed} ($\mathcal{M}_{q}^{\text{Mixed}}$).
Mixed memories represent a composite form of structural memory, combining all the aforementioned types: chunks, knowledge triples, atomic facts, and summaries. 
This integration provides a comprehensive representation, formally defined as follows: $\mathcal{M}_{q}^{\text{Mixed}} = \mathcal{C}_q \cup \mathcal{T}_q \cup \mathcal{A}_q \cup \mathcal{S}_q$.
% \begin{align}
%     \mathcal{M}_{q}^{\text{Mixed}} = \mathcal{C}_q \cup \mathcal{T}_q \cup \mathcal{A}_q \cup \mathcal{S}_q \, .
% \end{align}

% \begin{tcolorbox}[colback=gray!5!white,colframe=gray!60!black,title=Mixed Memory]
% \textbf{Definition:} Mixed memory combines all types of structural memories—chunks, knowledge triples, atomic facts, and summaries—providing a comprehensive representation. \\
% \textbf{Example:} From the same document:
% \begin{itemize}
%     \item \textbf{Chunk:} \textit{"In 1687, Sir Isaac Newton published his groundbreaking work 'Principia,' where he formulated the laws of motion."}
%     \item \textbf{Triple:} $\langle$Isaac Newton, published, Principia$\rangle$
%     \item \textbf{Atomic Fact:} "Principia formulated the laws of motion and universal gravitation."
%     \item \textbf{Summary:} \textit{"Newton's 'Principia' introduced the laws of motion and universal gravitation, foundational for classical mechanics."}
% \end{itemize}
% \end{tcolorbox}

% By leveraging mixed memories, agents benefit from a holistic view of the context, ensuring that no critical information is overlooked.
%
Details of the prompts used by the LLM for generating each type of structural memory, e.g., $\mathcal{P}_\mathcal{T}$, $\mathcal{P}_\mathcal{A}$ and $\mathcal{P}_\mathcal{S}$, are provided in Appendix~\ref{appendix:prompts}.
% Our empirical findings show that mixed memory consistently delivers balanced and competitive performance by integrating the complementary strengths of different memory structures. 
% Chunks and summaries excel in tasks involving extensive and lengthy contexts by ensuring coherence and continuity. 
% In contrast, knowledge triples and atomic facts capture relational and precision information, making them particularly effective for tasks requiring multi-hop reasoning and factual accuracy.
% Once generated, the structural memories are encoded into vector representations using an embedding model and subsequently stored in a vector database~\cite{douze2024faiss}. 
% This storage mechanism facilitates efficient similarity searches, which play a crucial role during the memory retrieval phase, as elaborated in the following section.

\subsection{Memory Retrieval Methods}
\label{subsec:memory_retrieval_methods}
Given the generated structural memories $\mathcal{M}_q$, we employ a memory retrieval method to identify and integrate the most relevant supporting memories $\mathcal{M}_r \subset \mathcal{M}_q$ for the query $q$.
Without this step, the agent would need to process all available memories, leading to inefficiency and potential inaccuracies due to irrelevant information. 
Our study mainly focuses on three retrieval approaches: single-step retrieval~\cite{robertson2009probabilistic, rubin2022learning}, reranking~\cite{gao2023precise, ji2024dynamic}, and iterative retrieval~\cite{li2024corpuslm, shi2024generate}. 
The details of each memory retrieval method are outlined as follows:

\noindent \textbf{Single-step Retrieval}. 
In the single-step retrieval process, the goal is to identify the Top-$K$ memories $\mathcal{M}_r$ that are most relevant to the query $q$.
This process is formally defined as: $\mathcal{M}_r = \texttt{Retriever}(q, \mathcal{M}_q, K)$, 
% \begin{align}
%     p(\mathcal{M}_r | q, \mathcal{M}_q) &= p_{\text{Retriever}}(\mathcal{M}_r | q, \mathcal{M}_q, K) \, ,
% \end{align}
% \begin{align}
%     \mathcal{M}_r = \texttt{Retriever}(q, \mathcal{M}_q, K) \, ,
% \end{align}
where the Retriever~\cite{robertson2009probabilistic, rubin2022learning} serves as the core component. 
% Examples of widely used Retriever include BM25~\cite{robertson2009probabilistic}, SBERT~\cite{reimers2019sentence}, and Dense Retrieval~\cite{rubin2022learning}.
% It encompasses three primary functionalities: encoding memories in high-dimensional embeddings, indexing for efficient search, and maintaining a memory datastore in the form of key value pairs~\cite{wu2024retrieval}.

\noindent \textbf{Reranking}.
% reranking introduces an additional refinement step aimed at enhancing the quality of the selected items.
In the reranking process~\cite{gao2023precise, dong2024don}, an initial retriever selects a candidate set of Top-$K$ memories $\mathcal{M}_i$, which are then reranked by an LLM prompted with $\mathcal{P}_{\text{Rerank}}$ based on their relevance scores. 
From this reranked list, the Top-$R$ memories $\mathcal{M}_r$, selected in descending order of relevance scores, are identified as the most relevant. 
This step enhances retrieval precision by leveraging the LLM to strengthen query-memory connections, filtering out irrelevant memories, and prioritizing the most pertinent memories for the query.
This process is formally defined as: $\mathcal{M}_r = \texttt{LLM}(q, \mathcal{M}_i, R, \mathcal{P}_R)\,$, where $\mathcal{M}_i = \texttt{Retriever}(q, \mathcal{M}_q, K)$.
% ~\footnote{For simplicity, we omit the inputs of the Top-$K$ retriever, including $q, \mathcal{M}_q, K$.}:
% \begin{align}
%     &p(\mathcal{M}_r | q, \mathcal{M}_q) \\
%     &= p_{\text{LLM}}(\mathcal{M}_r | q, \mathcal{M}_i, R, \mathcal{P}_R) \cdot p_{\text{Retriever}}(\mathcal{M}_i | \cdot) \, , \notag
% \end{align}
% \begin{align}
%     \mathcal{M}_r &= \texttt{LLM}(q, \mathcal{M}_i, R, \mathcal{P}_R)\, , \\
%     \mathrm{where} \ \ \mathcal{M}_i &= \texttt{Retriever}(q, \mathcal{M}_q, K) \, .
% \end{align}
%where the detail of prompt $\mathcal{P}_{\text{Rerank}}$ is provide in Appendix~\ref{appendix:prompts}.
% By integrating the coarse-grained retrieval capabilities of a traditional retriever with the nuanced reasoning capabilities of an LLM, reranking achieves a more refined selection of structural memories.

\noindent \textbf{Iterative Retrieval}.
The iterative retrieval approach~\cite{gao2023retrieval} begins with an initial query $q_0= q$ and retrieves the Top-$T$ most relevant structural memories $\mathcal{M}_j$.
These retrieved memories are used to refine the query through an LLM prompted by $\mathcal{P}_{\text{Refine}}$. 
This process is repeated over $N$ iterations, refining the query to produce the final version $q_N$ that is informative for retrieving relevant memories.
Formally, the iterative retrieval process can be defined as follows: $q_j = \texttt{LLM}(\mathcal{M}_{j}, \mathcal{P}_{\text{Refine}})$, where $\mathcal{M}_j = \texttt{Retriever}(q_{j-1}, \mathcal{M}_q, T)$.
% ~\footnote{For simplicity, we omit the inputs of the Top-$T$ retriever, including $q_{j-1}$ and $\mathcal{M}_q$.}:
% \begin{align}
%     &p(q_N | q) = \\
%     &\prod_{j=1}^{N} p_{\text{LLM}}(q_j, |\mathcal{M}_{j}, \mathcal{P}_{\text{Refine}}) \cdot p_{\text{Retriever}}(\mathcal{M}_j |T, \cdot) \, . \notag
% \end{align}
% \begin{align}
%     q_j &= \texttt{LLM}(\mathcal{M}_{j}, \mathcal{P}_{\text{Refine}}) \, , \\
%     \mathrm{where} \ \ \mathcal{M}_j &= \texttt{Retriever}(\mathcal{M}_j |q_{j-1}, \mathcal{M}_q, T)\, . \notag
% \end{align}
After $N$ iterations, the final refined query $q_N$ is used to retrieve the Top-$K$ most relevant memories for answer generation. This step can be expressed as: $\mathcal{M}_r = \texttt{Retriever}(q_N, \mathcal{M}_q, K)$.
% \begin{align}
%     p(\mathcal{M}_r | q_N, \mathcal{M}_q) &= p_{\text{Retriever}}(\mathcal{M}_r | q_N, \mathcal{M}_q, K) \, .
% \end{align}
% \begin{align}
%     \mathcal{M}_r = \texttt{Retriever}(q_N, \mathcal{M}_q, K) \, .
% \end{align}
The detailed prompts $\mathcal{P}_{\text{Rerank}}$ and $\mathcal{P}_{\text{Refine}}$ can be found in Appendix~\ref{appendix:prompts}.
% Extensive experiments demonstrate that iterative retrieval emerges as the most effective method, outperforming the other two approaches and consistently achieving superior performance across various tasks.

\subsection{Answer Generation}
\label{subsec:answer_generations}
Finally, the agent leverages the LLM to generate the answer based on the retrieved memory.
To achieve this, we propose two methods of answer generation.
In the first method, termed \textit{Memory-Only}, the retrieved memories $\mathcal{M}_r$ are directly utilized as the context for generating the answer.
The second method, termed \textit{Memory-Doc}, uses the retrieved memories to locate their corresponding original documents from $\mathcal{D}_q$. 
These documents then serve as the context for answer generation, providing the agent with more detailed and contextually enriched information.
% Our findings suggest that \textit{Memory-Doc} is more effective for tasks with long contexts requiring continuity, such as reading comprehension, while \textit{Memory-Only} excels in tasks prioritizing precision, such as multi-hop QA.

\begin{table*}
    \begin{center}
        \resizebox{0.99\textwidth}{!}{
            \begin{tabular}{lccccccccccc}
                \toprule
                \multirow{2}{*}{\textbf{Memory Structure}} & \multicolumn{2}{c}{\textbf{HotPotQA}} & \multicolumn{2}{c}{\textbf{2WikiMultihopQA}} & \multicolumn{2}{c}{\textbf{MuSiQue}} & \multicolumn{2}{c}{\textbf{NarrativeQA}} & \multicolumn{2}{c}{\textbf{LoCoMo}} & \multicolumn{1}{c}{\textbf{QuALITY}} \\
                \cmidrule(lr){2-3} \cmidrule(lr){4-5} \cmidrule(lr){6-7} \cmidrule(lr){8-9} \cmidrule(lr){10-11} \cmidrule(lr){12-12} 
                & \textbf{EM} & \textbf{F1} & \textbf{EM} & \textbf{F1} & \textbf{EM} & \textbf{F1} & \textbf{EM} & \textbf{F1} & \textbf{EM} & \textbf{F1} & \textbf{ACC}\\
                \midrule
                Full Content & 55.50 & 75.77 & 44.00 & 54.33 & 36.00 & 51.60 & 7.00 & 24.99 & 13.61 & 41.82 & 81.50 \\
                \midrule
                \rowcolor{gray!20}
                \multicolumn{12}{c}{\textit{Single-step Retrieval}} \\ % ====== Single Retrieval ===========
                \midrule
                % ====== Chunk ===========
                \rowcolor{white!10} Chunks                       & \underline{61.50}     & 76.93     & 43.50 & 59.17     & \textbf{35.50} & \textbf{54.45}     & 13.50 & 29.78 & 9.95 & 40.63  & \underline{76.00} \\
                % ====== Triple =========== 
                \rowcolor{white!10} Triples                      & 59.50     & 74.09     & \underline{44.50} & \underline{60.82}     & 31.00 & 50.13     & 11.50 & 22.04 & 8.42 & 41.08  & 61.50 \\
                % ====== Atomic Fact ===========
                \rowcolor{white!10} Atomic Facts                 & \textbf{62.50}     & \textbf{77.22}     & 39.50 & 58.63     & 30.50 & 51.31     & 13.50 & 27.49 & 9.42 & 42.92  & 71.50 \\
                % ====== Summary ===========
                \rowcolor{white!10} Summaries                     & 57.00     & 74.81     & 42.00 & 57.21     & \underline{34.00} & \underline{52.83}     & \textbf{16.50} & \textbf{ 32.93} & \textbf{10.99} & \textbf{44.94} & \underline{76.00} \\
                \rowcolor{white!10} Mixed                   & 60.00     & \underline{77.10}     & \textbf{48.50} & \textbf{65.25}     & 33.00 & 51.65     & \underline{14.50} & \underline{29.86} & \underline{10.47} & \underline{44.73} & \textbf{78.00} \\ 
                % \rowcolor{white!10} Mix~(Chunk + Atomic Facts)  & \textbf{62.50}    & \textbf{77.32}     & 40.00 & 58.91     & 31.50 & 52.84     & 13.50 & 30.05 & \textbf{11.52} & \textbf{45.74} & \textbf{78.50} \\ 
                \midrule
                \rowcolor{gray!20} 
                \multicolumn{12}{c}{\textit{Reranking}} \\ % ====== reranking ===========
                \midrule
                % ====== Chunk ===========
                \rowcolor{white!10} Chunks                       & \underline{63.00} & 77.35  & \underline{45.00} & \underline{61.31}   & \textbf{37.00} & \textbf{55.32}  & \textbf{16.00} & \underline{31.63}   & \underline{9.95} & 43.47  & \textbf{78.50} \\
                % ====== Triple =========== 
                \rowcolor{white!10} Triples                      & 61.00 & 76.75  & 43.50 & 55.43   & 26.50 & 42.05  & 10.00 & 20.65   & 8.83 & 41.82  & 60.00 \\
                % ====== Atomic Fact ===========
                \rowcolor{white!10} Atomic Facts                 & \underline{63.00} & \underline{78.31}  & 40.50 & 59.31   & 28.50 & 49.95  & \underline{14.00} & 28.19   & 8.90 & 44.27  & 67.50 \\
                % ====== Summary ===========
                \rowcolor{white!10} Summaries                     & 61.00 & 77.80     & \underline{45.00} & 61.18   & \underline{35.50} & \underline{54.59}   & \textbf{16.00} & \textbf{32.26}   & \textbf{12.04} & \textbf{44.83} & 75.00 \\
                \rowcolor{white!10} Mixed                   & \textbf{65.00} & \textbf{78.58}     & \textbf{45.50} & \textbf{61.77} & 34.00 & 52.45 & 11.98 & 28.02 & 9.42 & \underline{44.51} & \underline{77.50} \\ 
                % \rowcolor{white!10} Mix~(Chunk + Atomic Facts)  & 59.50 & 75.77     & 43.50 & \textbf{62.08} & 29.50 & 49.40 & 10.23 & 24.55 & 8.38 & 43.33 & 75.50 \\ 
                \rowcolor{gray!20} 
                \midrule
                \multicolumn{12}{c}{\textit{Iterative Retrieval}} \\  % ====== Iterative Retrieval ===========
                \midrule
                % ====== Chunk ===========
                \rowcolor{white!10} Chunks                       & 63.00 & 79.10 & 46.50 & 62.13 & 37.00 & 56.78 & \underline{14.50} & \underline{30.88} & \underline{10.47} & \underline{45.14} & \underline{77.00} \\
                % ====== Triple =========== 
                \rowcolor{white!10} Triples                      & 64.00 & 78.78 & \underline{47.50} & 62.06 & \underline{38.00} & 55.93 & 10.50 & 21.67 & 9.47 & 41.41 & 60.50 \\
                % ====== Atomic Fact ===========
                \rowcolor{white!10} Atomic Facts                 & \underline{65.50} & \underline{81.29} & 44.00 & \underline{63.89} & 34.50 & \underline{57.55} & \underline{14.50} & 28.28 & 9.95 & 43.62 & 67.50 \\
                % ====== Summary ===========
                \rowcolor{white!10} Summaries                     & 60.50 & 78.11 & 46.50 & 62.35 & 33.50 & 53.12 & \textbf{17.00} & \textbf{31.79} & \textbf{12.04} & 43.93 & 75.00 \\
                \rowcolor{white!10} Mixed                   & \textbf{67.00} & \textbf{82.11} & \textbf{51.00} & \textbf{68.15} & \textbf{39.00} & \textbf{61.38} & 12.50 & 28.36 & 7.85 & \textbf{45.25} & \textbf{79.50} \\ 
                % \rowcolor{white!10} Mix~(Chunk + Atmoic Facts)  & \textbf{67.00} & 80.97 & 45.50 & 64.75 & \textbf{39.00} & \textbf{63.15} & 14.00 & 29.80 & 8.90 & \textbf{46.64} & 79.00 \\ 
                \bottomrule
            \end{tabular}
        }
    \end{center}
    \vspace{-1.0em}
    \caption{Overall Performance (\%) of various memory structures utilizing different retrieval methods across six datasets. The best performance is marked in boldface, while the second-best performance is underlined.}
    \label{table:RQ1_RQ2}
\end{table*}

\section{Experiments}

% \subsection{Research Questions.}

% \noindent \textbf{RQ1}: How do different structural memories affect the performance across various tasks?

% \noindent \textbf{RQ2}: How do different memory retrieval methods influence performance on various tasks?

% \noindent \textbf{RQ3}: How does performance compare between \textit{Memory-Only} and \textit{Memory-Doc}?

% \noindent \textbf{RQ4}: How do the hyperparameters of memory reasoning methods impact performance across tasks?

% \noindent \textbf{RQ5}: How robust are different structural memories in tasks involving noise documents?

\subsection{Datasets.}

We conduct experiments on six datasets across four tasks. For multi-hop long-context QA datasets, we experiment with HotPotQA~\cite{yang2018hotpotqa}, 2WikiMultihopQA~\cite{ho2020constructing}, and MuSiQue~\cite{trivedi2022musique}. The single-hop long-context QA task is evaluated with NarrativeQA~\cite{kovcisky2018narrativeqa} from Longbench~\cite{bai2023longbench}. Additionally, we leverage the LoCoMo dataset~\cite{maharana-etal-2024-evaluating} for dialogue-based long-context QA task, while the QuALITY~\cite{pang2022quality} dataset is used for the reading comprehension QA task\footnote{More details and statistics about the datasets are provided in Appendix~\ref{appendix:datasets}.}.

\subsection{Evaluation.}
To evaluate QA performance, we follow previous work \cite{li2024graphreader} and use standard metrics such as Exact Match (EM) score and F1 score for the datasets HotPotQA, 2WikiMultihopQA, MuSiQue, NarrativeQA and LoCoMo. For QuALITY, we follow the approach in \cite{lee2024human} and use accuracy as the evaluation metric, with $25\%$ indicating chance performance.

\subsection{Implementation Details.}
In our experiments, we use GPT-4o-mini-128k with a temperature setting of 0.2. The input window is set to $4k$ tokens, while the maximum chunk size is up to $1k$ tokens.
For text embedding, we employ the text-embedding-3-small model~\footnote{\url{https://platform.openai.com/docs/guides/embeddings/}} from OpenAI and store the vectorized memories using LangChain~\cite{Chase_LangChain_2022}.
% Details on the structural memory generation prompts can be found in Appendix~\ref{appendix:prompts}.

\section{Results and Analysis}

% We provide our main experimental results in this section. 

\subsection{Impact of Memory Structures}
\label{subsec:impact_of_memory_structures}

% \begin{figure}[!t]
%     \centering
%     \includegraphics[width=0.99\columnwidth]{figs/case_study_mix_memory_SR.pdf}
%     \caption{Case Study (Mix Memory (All), Draft).}
%     \label{fig:structural_memory_case_study}
% \end{figure}

\begin{figure*}[!t]
    \centering
    \includegraphics[width=0.99\textwidth]{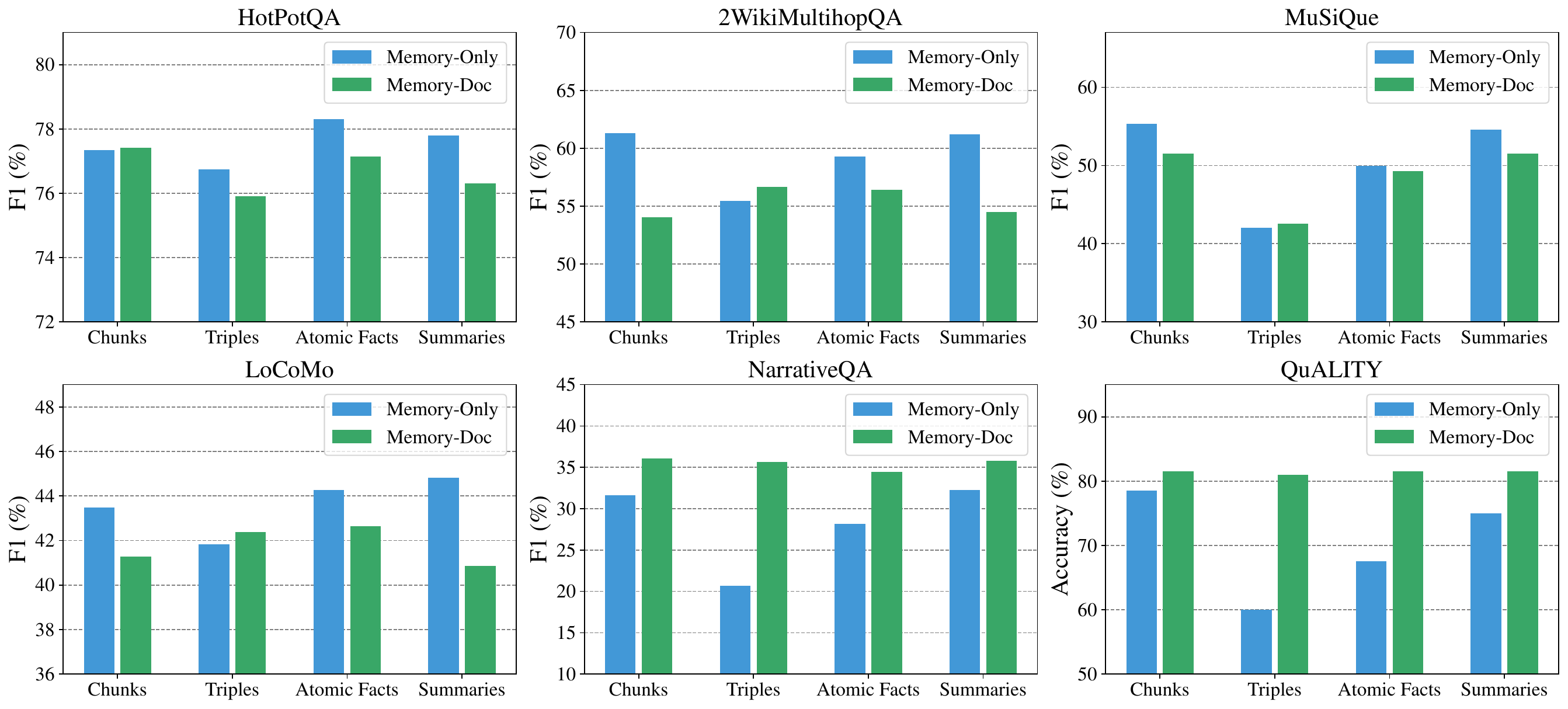}
    \caption{Performance across six datasets using two answer generation approaches: \textit{Memory-Only} and \textit{Memory-Doc}.}
    \label{fig:RQ-with-without-docs}
    \vspace{-1em}
\end{figure*}

\textbf{Finding 1: \textit{Mixed memories} delivers more balanced performance}. 
% To address \textbf{RQ1}, we conducted experiments on six datasets. 
The results as presented in Table~\ref{table:RQ1_RQ2} reveal key insights into the impact of various memory structures on task performance: 
(1) Mixed memories consistently outperform other memory structures.
This is particularly evident under iterative retrieval, where mixed memories achieve the highest F1 scores of 82.11\% on HotPotQA and 68.15\% on 2WikiMultihopQA.
(2) Chunks excel in tasks requiring a balance between concise and comprehensive contexts, as shown in datasets with long contexts. 
This is evidenced by its F1 score of 31.63\% on NarrativeQA and an accuracy of 78.5\% on QuALITY under reranking.
Summaries, which condense large contexts, is effective for tasks demanding abstraction, as shown by its competitive F1 score of 32.26\% on NarrativeQA and solid performance on LoCoMo.
% (3) Triples stand out for its ability to capture relational information, achieving an F1 score of 62.06\% on 2WikiMultihopQA under iterative retrieval. 
% Atomic facts excel in precision-oriented tasks, leading to F1 scores with 81.29\% on HotPotQA under iterative retrieval and performing consistently well on MuSiQue.
(3) Knowledge triples and atomic facts are particularly effective for relational reasoning and precision. Knowledge triples achieve an F1 score of 62.06\% on 2WikiMultihopQA under iterative retrieval, while atomic facts achieve an F1 score of 81.29\% on HotPotQA.
These findings emphasize the importance of tailoring memory structures to specific task requirements and demonstrate that integrating complementary memory types in mixed memories significantly enhances performance across tasks.

\subsection{Impact of Memory Retrieval Methods}
\label{subsec:impact_of_memory_retrieval}

\textbf{Finding 2: \textit{Iterative retrieval} as the optimal retrieval method}. 
% To answer \textbf{RQ2}, we evaluate three memory retrieval methods across six datasets. 
The results in Table~\ref{table:RQ1_RQ2} demonstrate the significant influence of the retrieval method on performance:
% , with iterative retrieval consistently outperforming the others:
(1) Iterative retrieval consistently outperforms the others, achieving the highest scores across most datasets.
Notably, with mixed memories, iterative retrieval achieved an F1 score of 82.11\% on HotPotQA and 68.15\% on 2WikiMultihopQA, showcasing its ability to refine queries iteratively for enhanced accuracy.
(2) Reranking demonstrates strong performance on datasets with moderate complexity. For instance, it achieved F1 scores of 44.27\% on LoCoMo and 28.19\% on NarrativeQA with atomic fact memory.
% , highlighting its effectiveness in optimizing precision by prioritizing relevant content.
(3) In contrast, single-step retrieval performs competitively in tasks requiring minimal contextual integration. 
Using summary memory, it achieved an F1 score of 32.93\% on NarrativeQA, leveraging abstraction to extract coherent information.
% Iterative retrieval stands out as the most effective strategy, consistently achieving the highest performance by refining queries through multiple iterations.
% For instance, when paired with atomic fact memory, it attained an F1 score of 82.29\% on HotPotQA
% This result underscores iterative retrieval’s capability to accumulate and refine information, thereby enhancing accuracy.
% (2) Reranking, by contrast, demonstrates robust performance on datasets of moderate complexity when coupled with specific memory types. For instance, with atomic fact memory, it achieved an F1 score of 44.27\% on LoCoMo and 28.19\% on NarrativeQA. These figures reflect reranking’s proficiency in enhancing retrieval precision by prioritizing relevant content.
% (3) Single-step retrieval, the most straightforward of the mechanisms, performs adequately in tasks with minimal contextual demands, particularly when paired with summary memory. On NarrativeQA, single-step retrieval achieved an F1 score of 32.93\%, leveraging the abstraction provided by summaries to extract coherent information. 
% Similarly, it demonstrated an accuracy of 78.0% on QuALITY with chunk-based memory, highlighting its suitability for simpler datasets.
% Overall, these insights emphasize the critical role of retrieval mechanisms in aligning QA performance with task complexity. 
% Iterative retrieval emerges as the most effective for tasks that demand intensive reasoning, while other mechanisms show promise in specific contexts.
These findings emphasize the importance of aligning retrieval mechanisms with task requirements, and iterative retrieval excels in reasoning tasks.
% These findings emphasize the pivotal role of iterative retrieval in maximizing the reasoning capabilities of LLM-based agents.
% , while reranking and single-step retrieval provide advantages in specific scenarios, depending on task complexity and memory type.

\subsection{Impact of Answer Generation Approaches}
\label{subsec:impact_of_with_or_without_docs}

\textbf{Finding 3: Extensive Context tasks favor \textit{Memory-Doc}, while precision tasks benefit from \textit{Memory-Only}.}
% To answer \textbf{RQ3}, we evaluate the \textit{Memory-Only} and \textit{Memory-Doc} approaches, as shown in Figure~\ref{fig:RQ-with-without-docs}, which compares their performance across various datasets.
As shown in Figure~\ref{fig:RQ-with-without-docs}, which compares their performance across various datasets.
% which generates answers directly from retrieved memories (w/o Docs), and the memory-document approach, which generates answers from documents retrieved via memories (w/ Docs). 
% Figure~\ref{fig:RQ-with-without-docs} illustrates the comparative analysis, shedding light on the performance of these methods across various datasets.
% In narrative-heavy and reading comprehension datasets, such as NarrativeQA and QuALITY, the \textit{Memory-Doc} approach outperforms.
% For instance, in NarrativeQA with triples-based memories, the \textit{Memory-Doc} achieves nearly double the performance of \textit{Memory-Only}. 
% The reason behind this is the critical role of contextual continuity in these datasets.
% By retrieving documents through retrieved memories, a more comprehensive understanding is facilitated, similar to how humans leverage both immediate memories and broader context to grasp complex narratives.
retrieving documents through retrieved memories provides a more comprehensive understanding, much like how humans integrate immediate recall with broader context to interpret complex narratives.
In contrast, for datasets involving multi-hop reasoning and dialogue understanding, such as HotPotQA and LoCoMo, the \textit{Memory-Only} approach proves to be the more effective strategy.
These findings highlight that tasks requiring extensive context benefit from the \textit{Memory-Doc} approach, which incorporates broader document-level information for enriched responses.
% The inclusion of external documents in these cases can introduce superfluous details, potentially detracting from the precision and focus required for accurate memory retrieval.
% These insights suggest that for tasks requiring a continuous and extensive context, a two-step process that begins with accessing retrieved memories and subsequently incorporates broader document-level information can lead to a more enriched and accurate response.
On the other hand, tasks prioritizing precision are better suited to the \textit{Memory-Only} approach, ensuring focused and accurate retrieval.
% This mirrors the human inclination to rely on precise memories for quick fact retrieval without the interference of unrelated contexts.

\begin{figure}[!t]
    \centering
    \includegraphics[width=0.99\columnwidth]{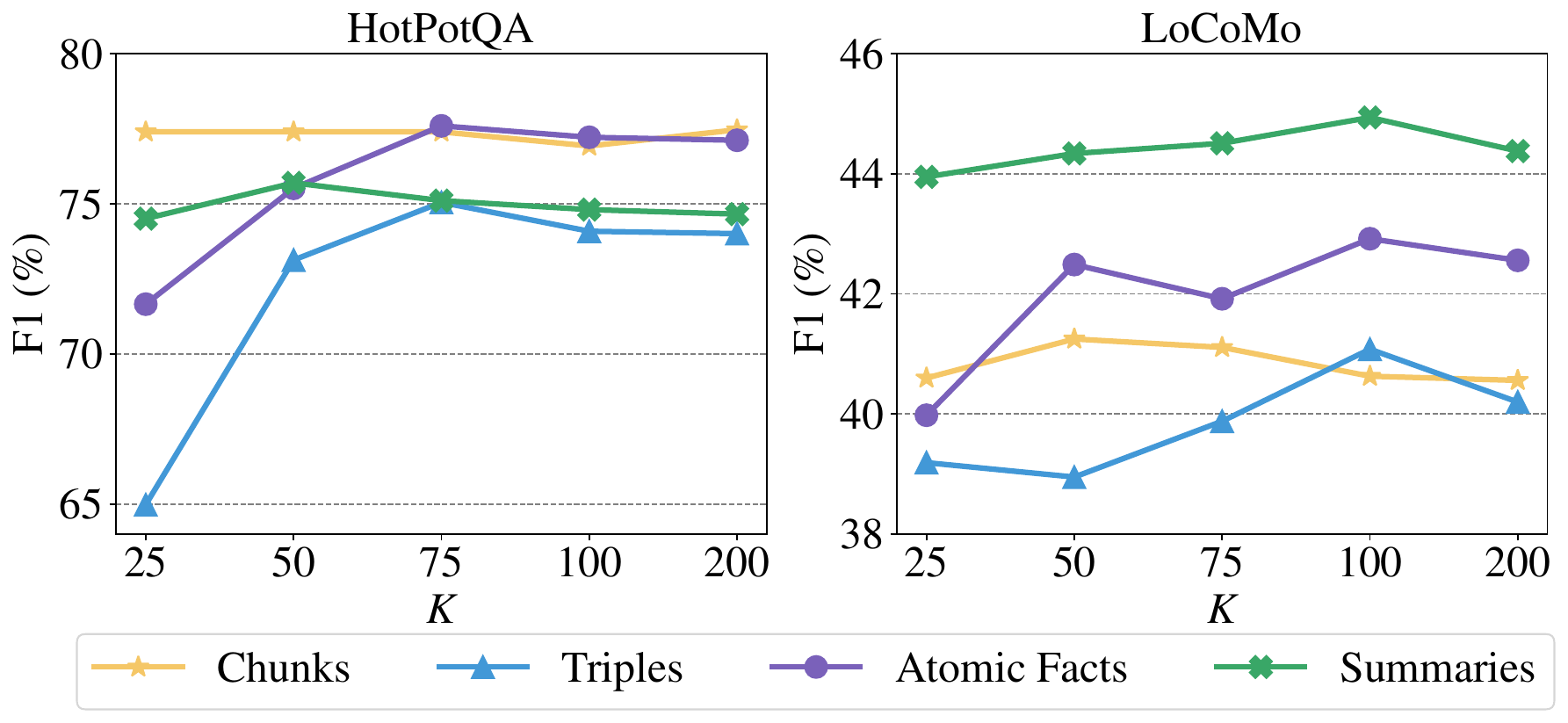}
    \caption{Performance of different numbers of retrieved memories $K$ on HotPotQA and LoCoMo using single-step retrieval.}
    \label{fig:RQ3-SR}
    % \vspace{-1em}
\end{figure}

\begin{figure}[!t]
    \centering
    \includegraphics[width=0.99\columnwidth]{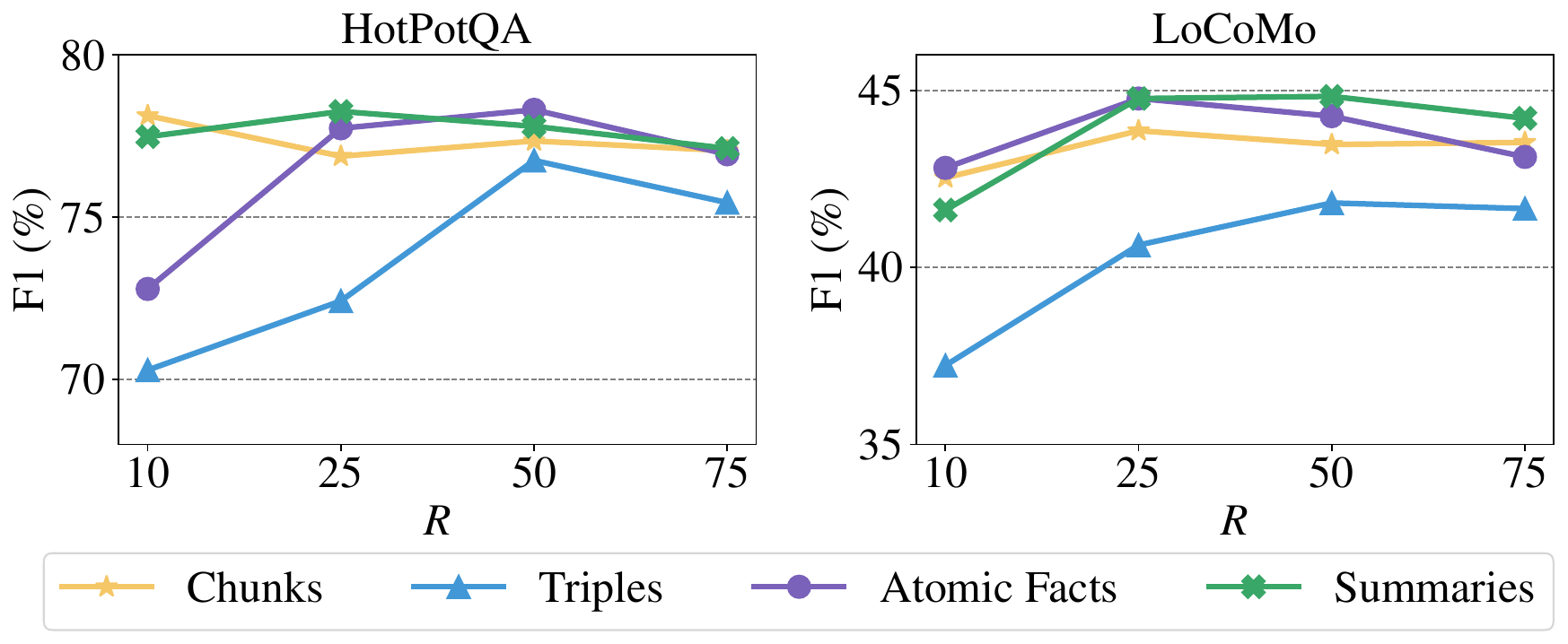}
    \caption{Performance of different numbers of reranked memories $R$ on HotPotQA and LoCoMo in reranking.}
    \label{fig:RQ3-RR}
\end{figure}

% \begin{figure*}[!t]
%     \centering
%     \includegraphics[width=0.99\textwidth]{figs/RQ3-IR.pdf}
%     \caption{F1 performance across varying numbers of memory units $T$ and iteration turns $N$ for two datasets (e.g., HotPotQA and LoCoMo) and four memory structures (e.g., chunk, triple, atomic fact, and summary) using iterative retrieval.}
%     \label{fig:RQ3-IR}
% \end{figure*}

\subsection{Hyperparameter Sensitivity}
\label{subsec:hyperparamter_sensitivity}
% We next address \textbf{RQ4}, investigating the hyperparameter sensitivity of memory retrieval methods.
% how the performance of QA tasks is influenced by the sensitivity of different memory structures, such as chunks, triples, atomic facts, and summaries, to the hyperparameters of their respective retrieval mechanisms. 
% These parameters include the number of retrieved memories $K$ used in single-step retrieval, the number of reranked memories $R$ in reranking, and both the number of retrieved memories $T$ and iteration turns $N$ in iterative retrieval. 
% To ensure comprehensive evaluation, we evaluate performance on HotPotQA and LoCoMo.

% \subsubsection{Impact of Number of Retrieved Memories $K$ on Single-step Retrieval Performance}

% \subsubsection{Impact of $K$}
\noindent \textbf{Effect of Number of Retrieved Memories $K$}.
% In the single-step retrieval, the Top-$K$ memories are retrieved based on the query to serve as the context for answer generation. 
We first evaluate the impact of $K$ in single-step retrieval, with a limit of $K=200$ due to computational resource limitations.
As depicted in Figure~\ref{fig:RQ3-SR}, in HotPotQA, chunks demonstrate consistent performance, stabilizing around 77\% across all $K$ values. 
% In contrast, the triples exhibit a significant increase, rising from 65\% at $K=25$ to approximately 75\% at $K=75$.
% This pattern is echoed in the atomic fact-based memory, which began at 72\% at $K=25$ and reached its peak performance at $K=75$.
% In LoCoMo, the influence of $K$ on memory structures was more nuanced.
In LoCoMo, the chunks show moderate gains up to $K=50$,
% showed a moderate increase from $K=25$ to $K=50$, after which it stabilized.
whereas triples, atomics, and summaries improve up to $K=100$ but then declined at $K=200$, likely due to noise introduced by retrieving excessive memories.
% This suggests that retrieving an excessive number of memory units may introduce noise, which can hinder performance in dialogue-oriented tasks.
% Overall, these results indicate that the optimal $K$ is contingent upon both the characteristics of the dataset and the specific memory structure in use. 
% While moderate $K$ values generally enhance performance, overly large $K$ values may introduce irrelevant information, thereby diminishing performance.
These findings indicate that the optimal $K$ depends on both the dataset and memory structure. 
While moderate $K$ values generally enhance performance, excessively large values can introduce irrelevant information, leading to a degraded performance.

% \subsubsection{Impact of Number of Reranked Memories $R$ on reranking Performance}
\noindent \textbf{Effect of Number of Reranked Memories $R$}.
% In reranking, the process begins by retrieving the Top-$K$ memories for a given query. the initial step involves retrieving the Top-$K$ memories based on a given query. 
% Following this, an LLM is used to rerank these memories, selecting the Top-$R$ memories to serve as the final context for answer generation. 
% In the reranking setup, the process begins by retrieving the Top-$K$ memories for a given query. 
% These are then reranked using an LLM, with the Top-$R$ memories selected as the final context for answer generation.
To evaluate the impact of $R$ in reranking, we investigate performance across a range of values, with a maximum $R$ of 75 due to computational cost constraints, while fixing $K$ at 100. 
As depicted in Figure~\ref{fig:RQ3-RR}, the results highlight that increasing the number of reranked memories does not always lead to better performance.
For instance, chunks achieve the highest F1 score at $R=10$ in HotPotQA, with a subsequent decline in performance beyond $R=50$.
This pattern is consistent with triples and atomic facts, indicating that selecting a smaller number of highly relevant memories can outperform retrieving and reranking larger sets, which often introduces noise.
% In LoCoMo, the reranking similarly demonstrates enhanced performance across memory structures.
A similar trend can be observed in LoCoMo.
% For example, the summaries reach peak performance at $R=50$ before experiencing a slight decline, while triples improve steadily from 36\% at $R=10$ to around 42\% at $R=75$.
These findings suggest that reranking is more effective when it focuses on a smaller subset of highly relevant memories. 
% Selecting an appropriate $R$ allows for optimal performance by balancing relevance and avoiding the diminishing returns and noise associated with larger selections.
% indicating that the top 50 reranked memory units provide sufficient context, with further units potentially introducing noise. 
% Similarly, the triple-based memory improves steadily from 36\% at $R=10$ to around 42\% at $R=75$.
% showing that reranking effectively enhances the context relevance.
% Overall, these findings indicate that reranking enables efficient retrieval by filtering out less relevant memory units, allowing to achieve high performance with moderate $R$ values while mitigating noise from larger selections.

\begin{figure}[!t]
    \centering
    \includegraphics[width=0.99\columnwidth]{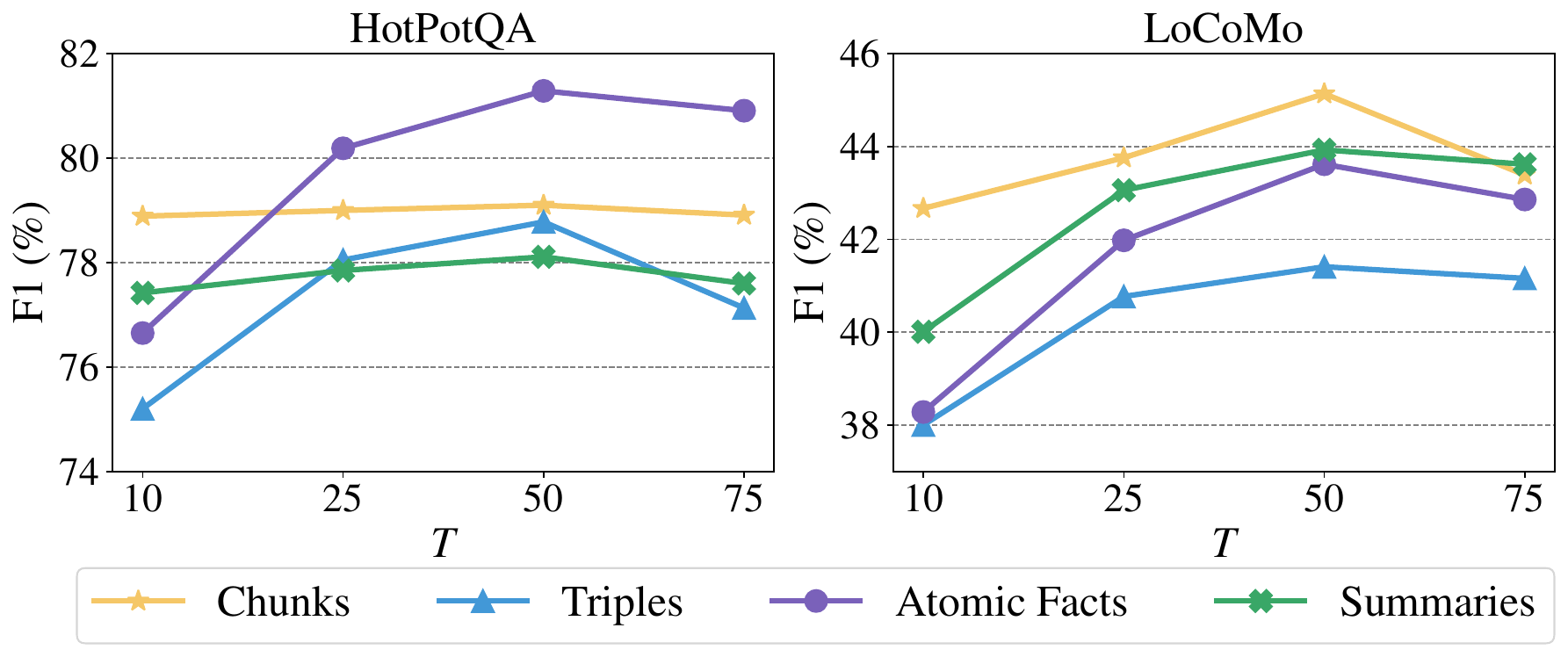}
    \caption{Performance of different numbers of retrieved memories $T$ in each interaction on HotPotQA and LoCoMo using iterative retrieval.}
    \label{fig:RQ3-IR-T}
    % \vspace{-1em}
\end{figure}

% Performance on HotPotQA and LoCoMo with different numbers of reranked memories. 

% \subsubsection{Impact of Memory Units $T$ and Iteration Turns $N$ on Iterative Retrieval Performance}
\noindent \textbf{Effect of Number of Retrieved Memories $T$ on Each Iteration}.
We first investigate performance across a range of values of $T$ using iterative retrieval, with a maximum $T$ of 75 and $N$ of 4 due to computational cost constraints while keeping $K$ fixed at 100. 
As illustrated in Figure~\ref{fig:RQ3-IR-T}, increasing the number of retrieved memories per iteration generally improves performance across datasets, though the gains diminish beyond a certain threshold.
For instance, in HotPotQA, atomic facts achieve an F1 score of approximately 81\% at $T=50$, with minimal additional gains from increasing $T$ further.
Similarly, in LoCoMo, chunks improve up to $T=50$ before declining at $T=75$.
These results indicate that while increasing $T$ can enhance query refinement and performance, excessively large $T$ values may introduce noise, ultimately reducing effectiveness.
% A similar trend is observed fro triples-based memory, which achieves around 78\% at $T=50$.
% In the LoCoMo dataset, chunk-based memory performance improves up to $T=50$ but declines at $T=75$.
% These findings suggest that while increasing $T$ can progressively refine the query and enhance performance, an excessively large $T$ may introduce irrelevant information, ultimately leading to a decrease in performance.

\begin{figure}[!t]
    \centering
    \includegraphics[width=0.99\columnwidth]{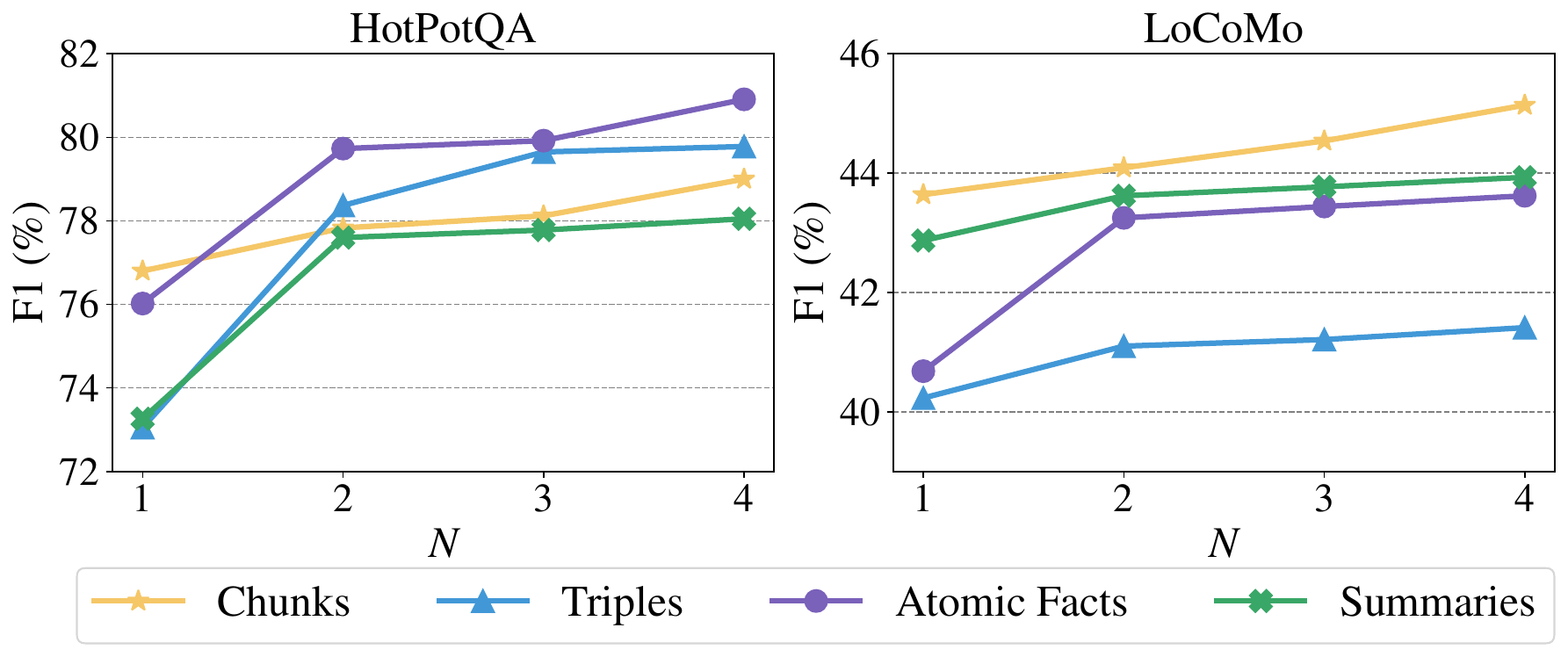}
    \caption{Performance of different numbers of retrieved memories $N$ in each interaction on HotPotQA and LoCoMo using iterative retrieval.}
    \label{fig:RQ3-IR-N}
    % \vspace{-1em}
\end{figure}

\noindent \textbf{Effect of Number of Iteration Turns $N$}.
Next, we examine the impact of iteration turns $N$, with the number of retrieved memories $T$ fixed at 50. 
As depicted in Figure~\ref{fig:RQ3-IR-T}, the results reveal that increasing $N$ initially enhances performance significantly, but the rate of improvement diminishes as $N$ continues to rise.
% Next, we investigate the impact of iteration turns $N$. we fix the retrieved memories $T$ at 50 at each iteration , we observe that an increase in $N$ initially leads to enhanced performance. 
% However, this improvement hover around at a point as $N$ continues to rise.
% For HotPotQA, both triple-based and summary-based memory structures show significant improvements from $N=1$ to $N=3$.
% Beyond this point, the gains are less pronounced. 
For HotPotQA, both triples and summars show notable gains from $N=1$ to $N=3$, after which the improvements become marginal.
% Similarly, chunk-based and atomic fact-based memories continue to benefit until $N=4$, after which the performance end to increase.
In the case of LoCoMo, triples, atomic facts, and summaries reach a peak at $N=3$ and stop increasing afterwards.
% the chunk-based memory structure demonstrates a consistent performance increase up to $N=4$, reaching a peak of approximately 45\%.
% In contrast, the performance of triple-based, atomic fact-based, and summary-based memories levels off around $N=3$.
% In LoCoMo, the chunk-based memory structure shows a steady rise in performance up to $N=4$, reaching approximately 45\%, while triple-based, atomic fact-based, and summary-based memories plateau around $N=3$. 
% These findings indicate that an intermediate number of iteration turns, typically between 2 and 3, strikes a balance between maximizing performance and avoiding excessive computational expenses.
These results suggest that an intermediate number of iteration turns, typically between 2 and 3, achieves optimal performance improvements, striking a balance between maximizing effectiveness and minimizing resource expenditure.

\begin{figure}[!t]
    \centering
    \includegraphics[width=0.99\columnwidth]{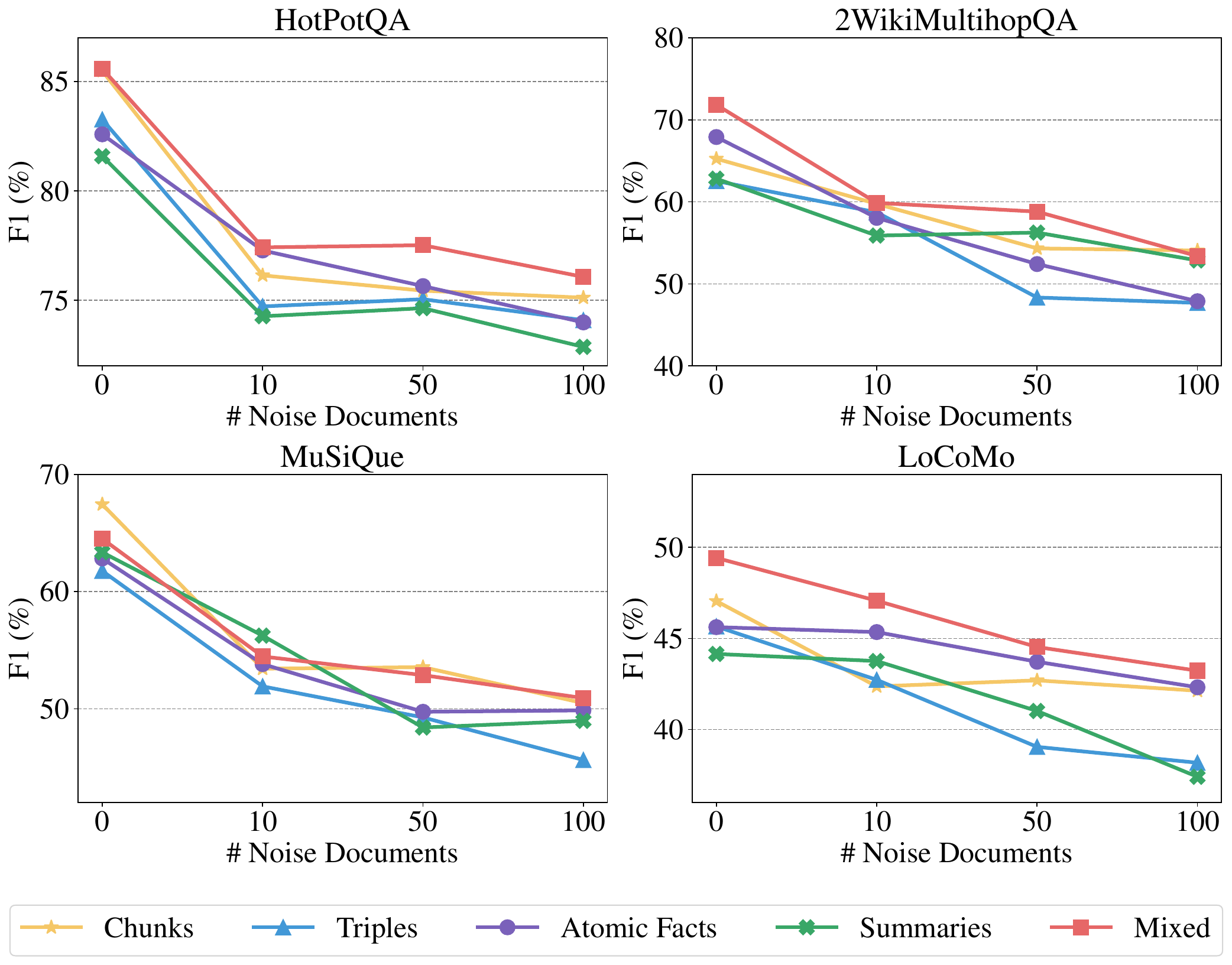}
    \caption{Performance across varying numbers of noise documents using single-step retrieval.}
    \label{fig:RQ5}
    % \vspace{-1em}
\end{figure}

\subsection{Impact of Noise Documents}
\label{subsec:impact_of_noise}
\textbf{Finding 4: \textit{Mix memory} excels in noise resilience.}
% Finally, to address \textbf{RQ5}, we evaluate the robustness of various memory structures, e.g., chunks, triples, atomic facts, summaries, and mixed memories, under increasing levels of noise.
Finally, we evaluate the robustness of various memory structures under increasing levels of noise using single-step retrieval with a fixed $K=100$.
% Specifically, we conduct experiments on two datasets (e.g., HotPotQA and LoCoMo) using single-step retrieval with a fixed $K=100$.
% As shown in Figure~\ref{fig:RQ5}, increasing the number of noise documents generally leads to a stark decline in the performance across all memory structures.
% For instance, in the case of HotPotQA, the F1 score of all memory structures generally decreases as the number of noise documents increases. 
% The mix memory structure performs the best throughout, while summary-based and triple-based memory exhibit relatively similar declines. 
% The chunk-based memory experiences a slower decline, maintaining a higher F1 score compared to others as the number of noise documents grows.
As depicted in Figure~\ref{fig:RQ5}, the performance of all memory structures declines as the number of noise documents increases.
For HotPotQA, the mix memory consistently achieves the highest F1 scores, demonstrating superior resilience to noise.
While triples and summaries exhibit similar rates of decline, the chunks experience a slower decline, maintaining a competitive F1 score when increasing the number of noise documents.
% A similar trend can be observed in LoCoMo. 
% Here again, the mix memory achieves the highest performance, and the chunk-based memory demonstrates a relatively stable decline. 
% These findings suggest that across all datasets, the mix memory structure consistently outperforms the others. 
A similar pattern is shown in LoCoMo.
% , where mixed memories achieve the best overall performance, and chunks exhibit notable stability against noise. 
These findings reveal the robustness of the mixed memory structure, which consistently outperforms others across datasets, making it the most effective choice in noisy environments.

\section{Conclusion \& Future Work}
In this paper, we present the first comprehensive study on the impact of structural memories and memory retrieval methods in LLM-based agents, 
% with the goal of identifying the most suitable memory structures for specific tasks and exploring how the choice of retrieval methods influence performance.
aiming to identify the most suitable memory structures for specific tasks and explore how retrieval methods influence performance.
This study yielded several key findings:
% Key findings reveal that 
(1) Mixed memories consistently deliver balanced performance. 
Chunks and summaries excel in tasks involving lengthy contexts, such as reading comprehension and dialogue understanding, while knowledge triples and atomic facts are effective for relational reasoning and precision in multi-hop and single-hop QA.
(2) Mixed memories also demonstrate remarkable resilience to noise.
(3) Iterative retrieval stands out as the most effective memory retrieval method, consistently outperforming in tasks such as multi-hop QA, dialogue understanding and reading comprehension.
While these findings provide valuable insights, further research is needed to explore how memory impacts areas such as self-evolution and social simulation, highlighting the importance of investigating how structural memories and retrieval techniques support these applications.
% Beyond reasoning, memory plays a critical role in enabling self-evolution as well as in action-oriented tasks, such as decision-making and goal-driven interactions. 
% Investigating how structural memories and retrieval techniques can support these advanced capabilities remains a crucial direction for future research.

\clearpage

\section*{Limitations}
We identify the following limitations in our work: 
(1) Our experiments are limited to tasks such as multi-hop QA, single-hop QA, dialogue understanding, and reading comprehension, which restricts the applicability of our findings to other complex domains like self-evolving agents or social simulation. Investigating the role of memory structures and retrieval methods in these topics could provide broader insights;
(2) The evaluation of memory robustness primarily considers random document noise, leaving other challenging noise types, such as irrelevant or contradictory information, unexplored. Investigating these addition noise in future studies could offer a more comprehensive understanding of memory resilience;
(3) Due to computational constraints, we limit the hyperparameter ranges (e.g., $K$, $R$, $T$, $N$) in memory retrieval methods. 
Expanding these ranges in future research could yield deeper insights into their impact on performance.

\bibliography{references}

\clearpage
\appendix

\section{Datasets}
\label{appendix:datasets}
We conduct experiments on the following six datasets across four tasks, including multi-hop QA, single-hop QA, dialogue understanding and reading comprehension.
The statistical information of datasets is provided in Table~\ref{table:dataset_statistic}.

\begin{table}[h]
    \begin{center}
        \resizebox{0.99\columnwidth}{!}{
            \begin{tabular}{lllcc}
                \toprule
                \textbf{Task} & \textbf{Dataset} & \textbf{Avg. \# Tokens} & \textbf{\# Samples} \\
                \midrule
                Multi-hop QA & HotpotQA & 1,362 & 200 \\
                Multi-hop QA & 2WikiMultihopQA &  985 & 200 \\
                Multi-hop QA & MuSiQue &  2,558 & 200 \\
                \midrule
                Single-hop QA & NarrativeQA &  24,009 & 200 \\
                \midrule
                Dialogue Understanding & LoCoMo & 24,375 & 191 \\
                \midrule
                Reading Comprehension & QuALITY &  4,696 & 200 \\    
                                                       % &                          &  & \\
                                                       % &                          &  & \\
                                                       % &                          &  & \\
                                                       % &                          &  & \\
                            
                \bottomrule
            \end{tabular}}
    \end{center}
    \caption{The statistic and example of datasets.}
    \label{table:dataset_statistic}
\end{table}

\section{Prompts}
\label{appendix:prompts}

In this section, we present the prompts employed in our experiments, with detailed descriptions provided in the respective subsections.

\subsection{Prompt for Generating Knowledge Triples}
\label{appendix:triple_prompts}
\begin{figure*}[tb]
    \centering
    \includegraphics[width=0.99\textwidth]{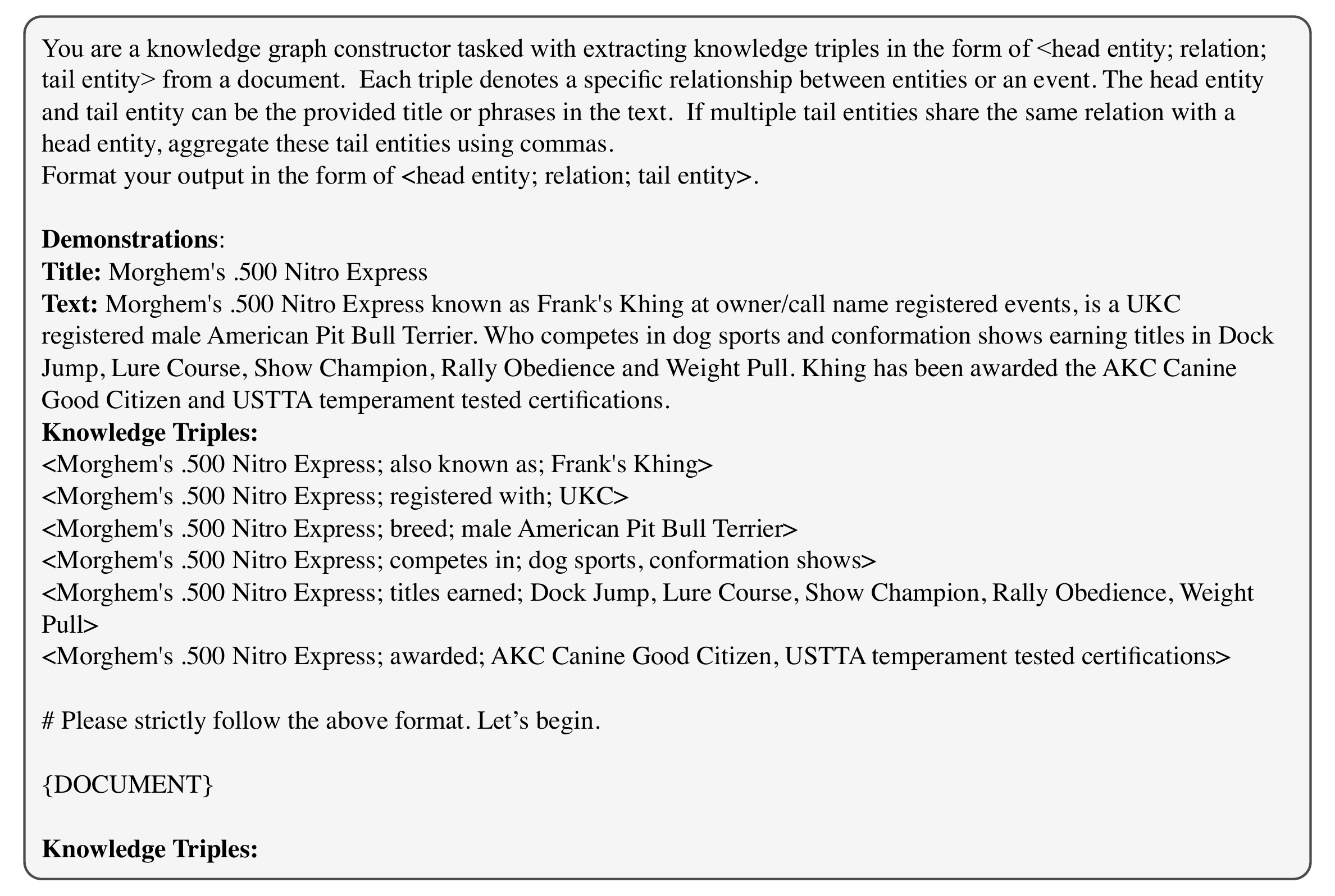}
    \caption{Prompt for generating knowledge triples from a document.}
    \label{fig:triple_prompt}
\end{figure*}

The prompt used for extracting knowledge triples from a document is illustrated in Figure~\ref{fig:triple_prompt}.

\subsection{Prompt for Generation Summaries}
\label{appendix:summaries_prompt}
\begin{figure*}[tb]
    \centering
    \includegraphics[width=0.99\textwidth]{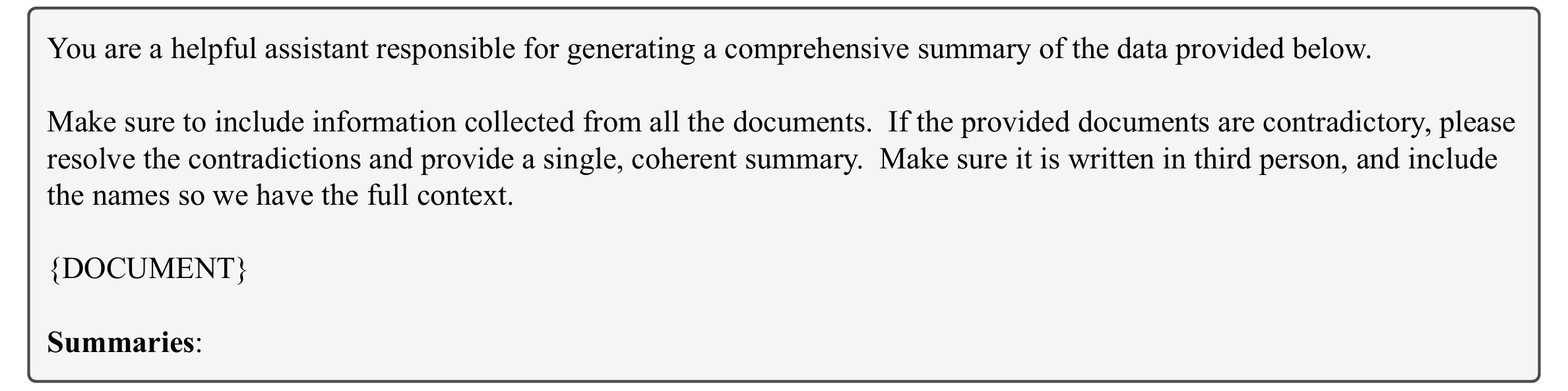}
    \caption{Prompt for generating summaries from a document.}
    \label{fig:summary_prompt}
\end{figure*}

The prompt designed for generating document summaries is depicted in Figure~\ref{fig:summary_prompt}.

\subsection{Prompt for Generating Atomic Facts}
\label{appendix:atomic_facts_prompt}
\begin{figure*}[tb]
    \centering
    \includegraphics[width=0.99\textwidth]{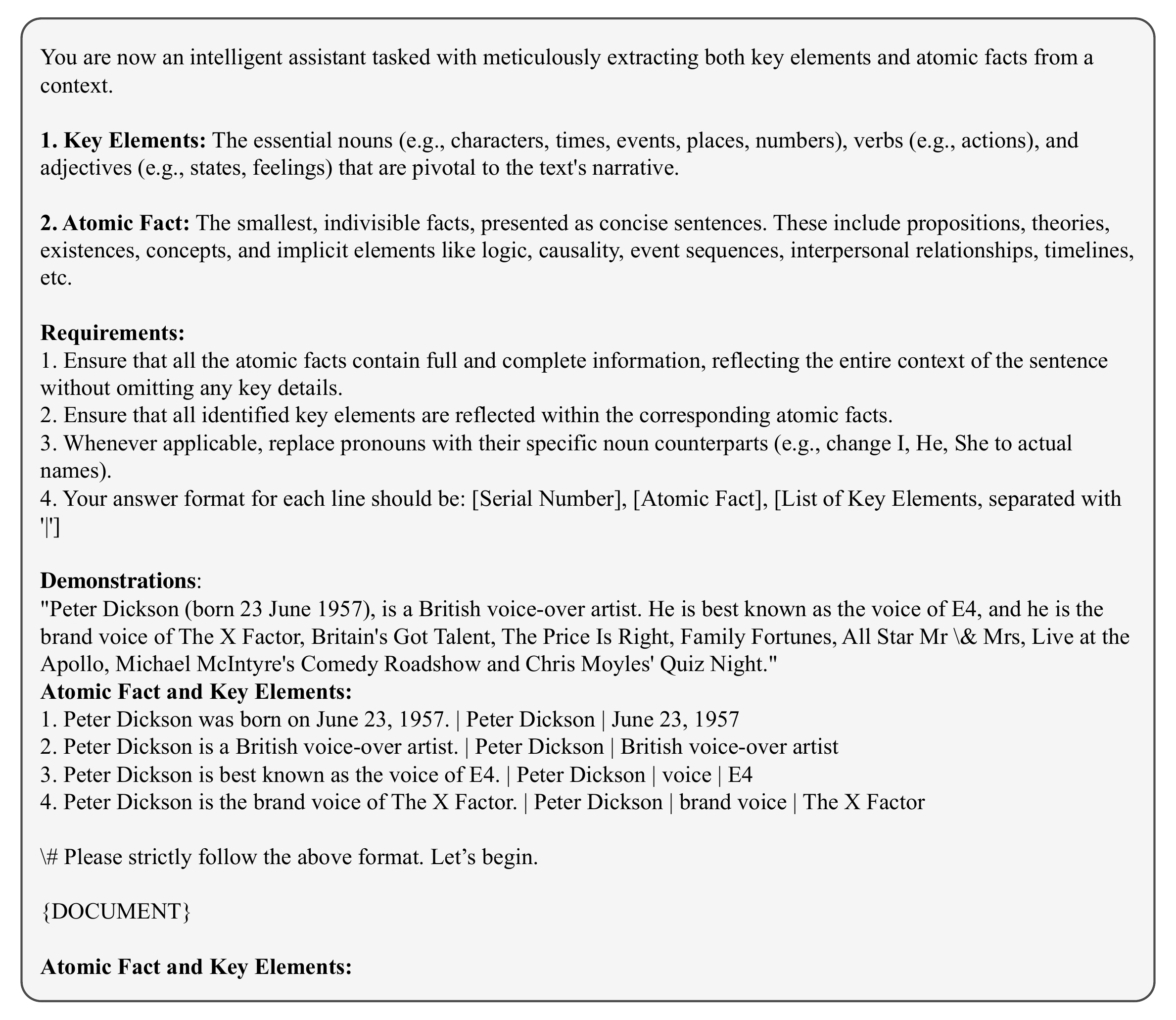}
    \caption{Prompt for generating atomic facts from a document.}
    \label{fig:atomic_fact_prompt}
\end{figure*}

The prompt for generating atomic facts from a document is shown in Figure~\ref{fig:atomic_fact_prompt}.

\subsection{Prompt for Reranking Retrieved Memories}
\label{appendix:reranking_prompt}
\begin{figure*}[tb]
    \centering
    \includegraphics[width=0.99\textwidth]{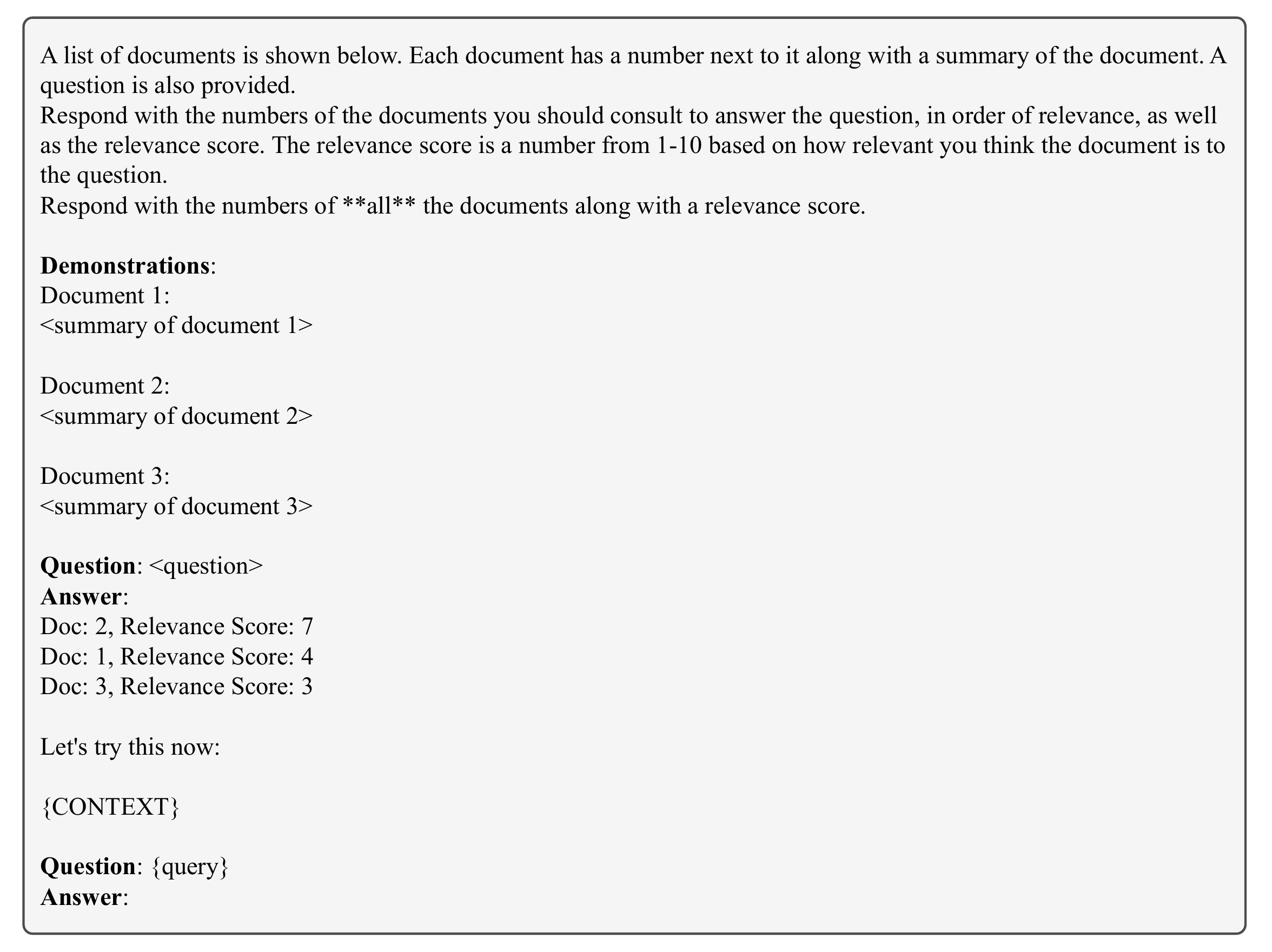}
    \caption{Prompt for reranking retrieved memories.}
    \label{fig:reranking_prompt}
\end{figure*}

The prompt used for reranking retrieved memories is presented in Figure~\ref{fig:reranking_prompt}.

\subsection{Prompt for Iterative Refining Query}
\label{appendxi:iterative_prompt}
\begin{figure*}[tb]
    \centering
    \includegraphics[width=0.99\textwidth]{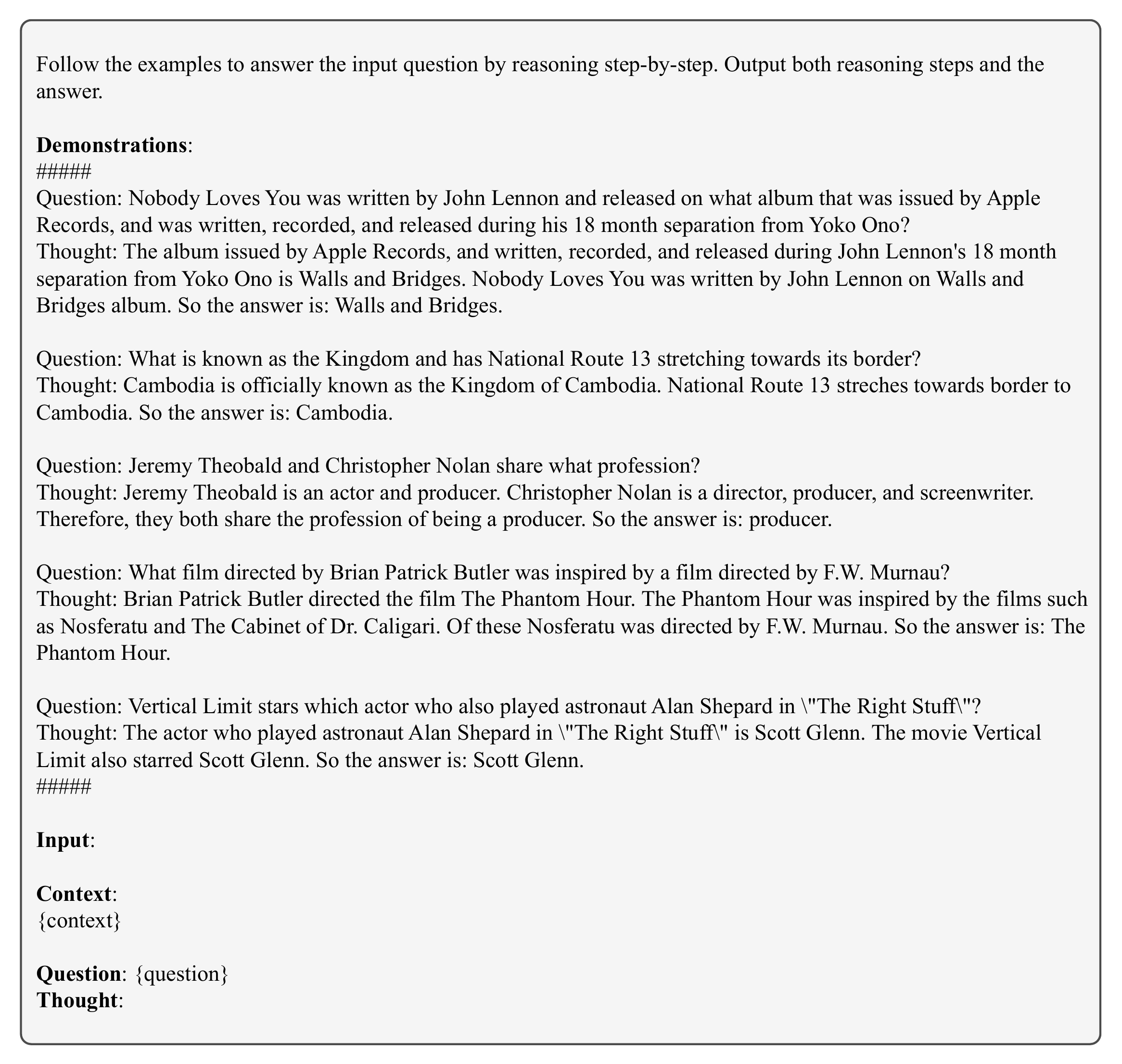}
    \caption{Prompt for the iterative refining query.}
    \label{fig:iterative_prompt}
\end{figure*}

The prompt for iterative query refinement is provided in Figure~\ref{fig:iterative_prompt}.

\end{document}